%% file: main.tex
\documentclass[11pt]{article}
\usepackage{amsmath,amsfonts,amsthm,amssymb,color, fullpage}
\usepackage{comment}
\usepackage{url}  
\usepackage{titlesec}

\newtheorem{theorem}{Theorem}[section]

\newtheorem{lemma}[theorem]{Lemma}

\usepackage{wrapfig}
\usepackage{subfig}
\usepackage{graphicx}
\theoremstyle{definition}
\newtheorem{definition}[theorem]{Definition}
\newtheorem{remark}[theorem]{Remark}

\newcommand{\R}{\mathbb{R}}

\usepackage{pdfpages}

\usepackage{thmtools, thm-restate}
\usepackage{algpseudocode}
\usepackage{algorithm}

\usepackage{xcolor}
\usepackage{graphicx}
\graphicspath{{images/}{../images/}}
\usepackage{subfiles}
\usepackage{natbib}

\newcommand{\Ex}{\mathbb{E}}

\DeclareMathOperator*{\argmax}{argmax}
\newcommand{\calA}{\mathcal{A}}
\newcommand{\batch}{\text{batch}}

\newcommand{\toRemove}[1]{}

\newcommand{\nmcomment}[1]{}

\title{Parallelizing Thompson Sampling}

\date{ }

\begin{document}

\author{
Amin Karbasi\\
Yale University\\
\and
Vahab Mirrokni \\ 
Google Research
\and 
Mohammad Shadravan\\
Yale University
}

\maketitle

\begin{abstract} 
\subfile{abstract}
\end{abstract}

\section{Introduction}
\input{introduction}

\section{Related Work}

\input{related}

\section{Preliminaries and Problem Formulation}
\label{sec:preliminaries}
\input{preliminaries}

\section{Batch Thompson Sampling for Stochastic Multi-armed Bandit}
\label{sec:alg}
\input{algorithm}

\label{sec:twoarm}

\input{twoarm}

\input{conclusion}

\bibliographystyle{plainnat}
\bibliography{mybib}

\onecolumn

\appendix
\section*{Appendices}
\label{sec:appendix}
\input{appendix}

\end{document}

%% file: abstract.tex
How can we make use of information parallelism in online decision making problems while efficiently balancing the exploration-exploitation trade-off? In this paper, we introduce a batch Thompson Sampling framework for two canonical online decision making problems, namely,  stochastic multi-arm bandit and linear contextual bandit with finitely many arms. Over a time horizon $T$,  our \textit{batch} Thompson Sampling policy achieves the same  (asymptotic) regret bound of a fully sequential one while carrying out only   $O(\log T)$ batch queries.  To achieve this exponential reduction, i.e., reducing the number of interactions from $T$ to $O(\log T)$, our batch policy dynamically determines the duration of each batch in order to balance the exploration-exploitation trade-off. We also demonstrate experimentally  that  dynamic batch allocation dramatically  outperforms natural baselines such as static batch allocations. 

%% file: introduction.tex
Many  problems in machine learning and artificial intelligence are sequential in nature and require making decisions over a long period of time and under uncertainty. Examples include  A/B testing \citep{graepel2010web}, hyper-parameter tuning \citep{kandasamy2018parallelised}, adaptive experimental design \citep{berry1985bandit}, ad placement \citep{schwartz2017customer}, clinical trials \citep{villar2015multi}, and recommender systems \citep{kawale2015efficient}, to name a few. Bandit problems provide a simple yet expressive view of sequential decision making with uncertainty. In such problems, a repeated  game between a learner and the environment is played where at each round the learner selects an action, so called an \textit{arm},  and then the environment reveals the reward. The goal of the learner is to maximize the accumulated reward over a  \textit{horizon} $T$. The main challenge faced by the learner is that the environment is unknown, and thus the learner has to follow a policy that identifies an efficient trade-off between the exploration (i.e., trying new actions) and exploitation (i.e., choosing among the known actions). A common way to measure the performance of a policy is through \textit{regret}, a game-theoretic notion, which is defined as   the difference between the reward accumulated by the policy and that of the best fixed  action in hindsight. 

We say that a policy has no regret, if its regret growth-rate as a function of $T$ is sub-linear. There has been a large body of work aiming to develop no-regret policies for a wide range of bandit problems (for a comprehensive overview, see \citep{lattimore2020bandit, bubeck2012regret, slivkins2019introduction}). However, almost all the existing policies are fully sequential in nature, meaning that once an action is executed the reward is immediately observed by the learner and can be incorporated to make the subsequent decisions. In practice however, it is often  more preferable (and sometimes the only way) to explore many actions in parallel, so called a \textit{batch of actions}, in order to gain more information about the environment in a timely fashion. For instance, in clinical trials, a phase of  medical treatment is often carried out on a group of individuals and the results are gathered for the entire group at the end of the phase. Based on the collected information, the treatment for the subsequent phases are devised \citep{batchedtwoarms}. Similarly, in a marketing campaign, the response to a line of products is not collected in a fully sequential manner, instead, a batch of products are mailed to a subset of costumers and their feedback is gathered collectively \citep{schwartz2017customer}.  Note that developing a no-regret policy is impossible without any information exchange about the carried out actions and obtained rewards. Thus, the main challenge in developing a batch policy is to balance between how  many actions to run in parallel  (i.e., batch size) versus how frequently to share information (i.e., number of batches). At  one  end of the spectrum lie the fully sequential no-regret bandit policies where the batch size is $1$, and the number of batches is $T$. At the other end of the spectrum lie the fully parallel policies where the  batch size is $T$ and all the actions are completely determined a priory without any amount of information exchange (such policies clearly suffer a linear regret).

In this paper, we investigate the sweet spot between the batch size and the  corresponding regret in the context of Thompson Sampling (TS). More precisely,

\begin{itemize}
    \item For the stochastic $N$-armed bandit, we develop Batch Thomson Sampling (B-TS), a batch version of the vanilla Thomson Sampling policy, that achieves the problem-dependent asymptotic optimal regret with $O(N\log T)$ batches. B-TS policy with the same number of batches also achieves the problem independent regret bound of $O(\sqrt{NT\log T})$ with Beta priors, and a slightly improved regret bound of $O(\sqrt{NT\log N})$ with Gaussian priors. 
    
    \item For the stochastic $N$-armed bandit, we  develop Batch Minimax Optimal Thompson Sampling (B-MOTS), a batch Thompson Sampling policy that achieves the optimal minimax problem-independent  regret bound of $O(\sqrt{NT})$ with  $O(N\log T)$ batches. We also present B-MOTS-J, a variant of B-MOTS, designed for Gaussian rewards, which  achieves  both  minimax  and  asymptotic optimality with $O(N \log(T))$ batches.
    \item Finally, for the linear contextual  bandit with $N$ arms, we develop Batch Thompson Sampling for Contextual Bandits (B-TS-C)  that achieves the problem-independent regret bound of $\tilde{O}(d^{3/2}\sqrt{T})$ with $O(N \log(T))$ batches.
\end{itemize}

The main idea that allows our batch policy to achieve near-optimal regret guarantees while reducing the number of sequential interactions with the environment from $T$ to $O(\log T)$ is a novel \textit{dynamic} batch mechanism that determines the duration of each batch based on an offline estimation of the regret accumulated during that phase.  We also observe empirically that  batch Thompson Sampling methods with a fixed batch size, but equal number of batches, incur higher regrets. 

%% file: related.tex
In this paper, we mainly focus on Thompson Sampling (also known as posterior sampling and probability matching), the earliest principled way for managing the exploration-exploitation trade-off in sequential decision making problems \citep{thompson1933likelihood, russo2017tutorial}. There has been a recent surge in understanding the theoretical guarantees of Thompson Sampling  due to its strong empirical evidence and  simple implementation \citep{chapelle2011empirical}. In particular, for the stochastic multi-armed bandit problem, \citet{shipra} proved a problem-dependent logarithmic  bound on expected regret of Thompson Sampling which was then showed to be asymptotically optimal \citep{kaufmann2012thompson}. Subsequently, \citet{agrawal2017near} provided  a problem-independent (i.e., worst-case) regret  bound of $O(\sqrt{NT\log T})$ on the expected regret when using Beta priors. Interestingly, the expected regret can be improved to  $O(\sqrt{NT\log N})$ by using Gaussian priors.   Very recently, \citet{MOTS} developed Minimax Optimal Thompson Sampling (MOTS), a variant of Thompson Sampling that achieves the minimax optimal regret of $O(\sqrt{NT})$.  \citet{contextual} also extended the analysis of multi-armed Thompson Sampling to the linear contextual setting and proved a  regret bound of $\tilde{O}(d^{3/2}\sqrt{T})$ where $d$ is the dimension of the context vectors.  In this paper, we develop the first variants of Batch Thompson Sampling  that achieve the aforementioned regret bounds (problem-dependent and problem-independent versions) while reducing the sequential interaction with the environment from $T$ to $O(N\log T)$, thus increasing the efficiency of running Thompson Sampling by an exponential factor (for a fixed $N$).

There has been a large body of work and numerous algorithms for regret minimization of multi-armed bandit problems, including upper confidence
bound (UCB), $\epsilon$-greedy, explore-then-commit, among many others. We refer the interested readers to some recent surveys for more details \citep{lattimore2020bandit, slivkins2019introduction}. The closest line of work to our paper is the   proposed batch UCB algorithm \citep{gao2019batched}, for which   \citet{esfandiari2021AAAI} showed an asymptotically optimal regret bound  with $O(\log T)$ number of batches. Very  recently,  \citet{esfandiari2021AAAI} and \citet{ruan2020linear} also addressed  the batch linear bandits and the batch linear contextual bandits, respectively. Our work extends those results to the case of Thompson Sampling for the stochastic multi-armed bandit as well as the linear contextual bandit problems. 

As we have highlighted in our proofs, our work builds on previous art, especially \cite{shipra, agrawal2017near, contextual} (we believe that giving due credits to previous work is a virtue and not vice). However, we build on a non-trivial way.  As it is clear from their analysis (and more generally for randomized probability matching strategies), breaking the sequential nature of distribution updates is non-trivial. We show that by a careful batch-mode strategy, one can reduce the sequential updates from $T$ to $O(\log(T))$. We are unaware of any previous work that obtains such a  result for Thompson Sampling. In contrast, UCB strategies are much more amenable to parallelization (and the analysis is simple) as one can simply use the arm elimination method proposed by \cite{esfandiari2021AAAI} and \cite{gu2021batched}. There is no clear way to use the arm elimination strategy for batch TS. Moreover, batch TS clearly outperforms the fully sequential UCB in all of our empirical results. 

The benefits of batch-mode optimization has been considered in other machine learning settings, including convex optimization \citep{balkanski2018parallelization, chen2019minimax}, submodular optimization \citep{chen2019unconstrained, fahrbach2019submodular, balkanski2018adaptive}, Gaussian processes \citep{desautels2014parallelizing, kathuria2016batched, contal2013parallel}, stochastic sequential optimization \citep{esf2021adaptivity, agarwal2019stochastic, chen2013near}, and Bayesian optimization \citep{wang2018batched, rolland2018high}, to name a few.

%% file: preliminaries.tex

As stated earlier, a standrad bandit problem is a repeated sequential game between a learner and the environment where at each round $t=1, 2, \dots, T,$ the learner selects an action $a(t)$ from the set of actions $\calA$ and then the environment reveals the reward $r_{a(t)}\in \R$. Different structures on the set of actions and rewards define different bandit problems. In this paper, we mainly consider two canonical variants, namely, stochastic multi-armed bandit, and stochastic linear contextual bandit.

\paragraph{Stochastic Multi-Armed  Bandit.}
In this setting, the set of actions $\calA$ is finite, namely, $\calA=[N]$, and each action $a\in[N]$ is associated with a sub-Gaussian distribution $P_a$ (e.g., Bernoulli distribution, distributions supported on $[0,1]$, etc). When the player selects an action $a$, a reward $r_{a}$ is sampled \textit{independently} from $P_{a}$. We denote by $\mu_a = \Ex_{a\sim P_a}[r_a]$ the average reward of an action $a$  and by $\mu^* =\max_{a\in \calA} \Ex_{a\sim P_a}[r_a]$ the action with the maximum average reward. Suppose the player selects actions $a_1,\dots,a_T$ and receives the stochastic rewards $r_{a(1)}, \dots, r_{a(T)}$. Then the (expected) regret is defined as 
\[
\mathcal{R}(T) =T\mu^*-\Ex \left[\sum_{t=1}^T r_{a(t)} \right].
\]
We say that a policy achieves \textit{no-regret}, if $\Ex \left[\mathcal{R}(T) \right]/T \rightarrow 0$ as the horizon $T$ tends to infinity. In order to compare the regret of algorithms, there are multiple choices in the literature. Once we fully specify the horizon $T$, the class of the bandit problem (e.g., multi-armed bandit with $N$ arms) and the specific instance we encounter withing the class (e.g., $\mu_1, \dots, \mu_N$ in the stochastic multi-armed problem), then we can consider the \textit{problem-dependent} regret bounds for each specific instance. In contrast, \textit{problem-independent} bounds (also called worst-case bounds) only depends on the horizon $T$ and class of bandits for which the algorithm is designed (which is the number of arms $N$ in the multi-armed stochastic bandit problem), and not the specific instance within that class. 
\footnote{There is a related  notion of regret, called \textit{Bayesian regret}, considered in the Thompson Sampling literature \citep{russo2014learning, bubeck2013prior}, where a known prior on the environment is assumed. The frequentist regret bounds considered in this paper immediately imply a regret bound on the Bayesian regret but the opposite is not generally possible \citep{lattimore2020bandit}.}
For the problem-dependent regret bound, it is known that UCB-like algorithms \citep{auer2002using, garivier2011kl, maillard2011finite} and Thomson Sampling \citep{agrawal2013further, kaufmann2012thompson} achieve the asymptotic regret of $O(\log T \sum_{\Delta_a>0} \Delta_a^{-1})$ where $\Delta_a= \mu^* - \mu_a \geq0$. It is also known that no algorithm can achieve a better asymptotic regret bound \citep{lai1985asymptotically}, thus implying that UCB and TS are both asymptotically optimal. In contrast, for the stochastic multi-armed bandit, UCB achieves the minimax problem-independent regret bound  of $\sqrt{NT}$ \citep{auer2002using} whereas TS (with Beta-priors) achieves a slightly worst regret of $\sqrt{NT \log T}$ \citep{agrawal2017near}. Very recently, \citet{MOTS} developed Minimax Optimal Thompson Sampling (MOTS) that achieves the minimax optimal regret of $O(\sqrt{NT})$.  


\paragraph{Contextual Linear Bandit.}

Contextual linear bandits generalise the multi-armed setting by allowing the learner to make use of side information. More specifically, each arm $a$ is associated with a feature/context vector $b_a\in \mathbb{R}^d$. At the beginning of each round $t\in[T]$,  the learner first observes the contexts $b_a(t)$ for all $a\in\mathcal{A}$, and then she chooses an action $a(t)\in\mathcal{A}$.  We assume that a feature vector $b_a$ affects the reward in a linear fashion, namely, $r_{a}(t) = \langle b_a(t), \mu \rangle + \eta_{a,t}$. Here, the parameter $\mu$ 
is unknown to the learner, and $\eta_{a,t}$ is an independent zero-mean sub-Gaussian noise given all the actions and rewards up to time $t$. Therefore, $\Ex[r_{a}(t)|b_a(t)] =\langle b_a(t), \mu \rangle$. The learner is trying to guess the correlation between $\mu$ and the contexts $b_a(t)$. For the set of actions $a(1),\dots,a(T)$, the regret is defined as 
\[
\mathcal{R}(T)= \left[ \sum_{t=1}^{T} r_{a^*(t)}(t)\right] -\left[ \sum_{t=1}^{T} r_{a(t)}(t)\right],
\]
where $a^*(t) = \arg\max_{a} \langle b_a(t) , \mu \rangle$. The context vectors at time $t$ are generally chosen by an adversary after observing the actions played and the rewards received  up to time $t-1$. In order to obtain scale-free regret bounds, it is commonly assumed that 
$\|\mu\|_2\leq1$
and
$\|b_a(t)\|_2\leq1$ for all arms $a\in\mathcal{A}$.
By applying UCB to linear bandit, it is possible to achieve $\mathcal{R}(T)= \tilde{O}(d \sqrt{T})$ with high probability \citep{auer2002using, dani2008stochastic, rusmevichientong2010linearly, abbasi2011improved}.
In contrast, \citet{contextual} showed that the regret of  Thompson Sampling can be bounded by $\tilde{O}(d^{3/2}\sqrt{T})$.

\paragraph{Batch Bandit.}
The focus of this paper is to parallelize the sequential decision making problem. In contrast to the fully sequential setting, where the learner selects an action and immediately receives the reward, in the batch mode setting, the learner selects a batch of actions and receives the rewards of all of them simultaneously (or only after the last action is executed). More formally, let the history $\mathcal{H}_{t}$ consists of all the actions and rewards  up to time $t$, namely,  $\{a(s)\}_{s\in[t-1]}$ and $\{r_{a(s)}(s)\}_{s\in[t-1]}$, respectively. We also denote the observed set of contexts up to and including time $t$ by  $C_t= \{b_{a}(s)\}_{a\in \mathcal{A}, s\in[t]}$. Note that in the multi-armed bandit problem $C_t=\emptyset$.   A fully sequential policy $\pi$ at round $t\in[T]$ maps the history and contexts to an action, namely, $\pi_t:\mathcal{H}_{t}\times C_t \rightarrow \mathcal{A}$.  In contrast, a batch policy $\pi$ only interacts with the environment at rounds $0=t_0<t_1<t_2 \dots <t_m=T$. The $l$-th batch of duration $t_{l} - t_{l-1}$ contains the time units $\{t_{l-1}+1,t_{l-1}+2, \dots, t_l\}$ which we denote it by the shorthand $(t_{l-1}, t_l]$. 
  To select the actions in the $l$-th batch the policy is only allowed to use the history of actions/rewards observed in the previous batches, in addition to the contexts received so far. Therefore, a batch policy at time $t\in(t_{l-1}, t_l]$ is the following map: $\pi_t:\mathcal{H}_{t_{l-1}}\times C_t \rightarrow \mathcal{A}$. Moreover, a batch policy with a predetermined fixed batch size is called \textit{static} and the one with a dynamic batch size is called \textit{dynamic}.

%% file: algorithm.tex
In the classic Thompson Sampling (TS), at any time $t\in[T]$, we consider a prior  distribution $D_{a}(t)$  on the underlying parameters of the reward distribution for every arm $a\in[N]$. TS works by  first sampling $\theta_a(t)\sim D_a(t)$, independently for each $a\in[N]$, and then choosing the one with the highest value, namely, $a_t = \argmax_{a\in[N]} \theta_a(t)$. Once the action $a_t$ is played, we receive the reward $r_t$, based on which the the prior distributions are updated as follows. If an arm $a$ is not selected, its distribution does not change, i.e., $D_a(t+1)=D_a(t)$. However, if $a=a_t$, then we update $D_a(t+1)$ given the information $(a_t, r_t)$ using the Bayes rule. By instantiating TS with different  prior distributions (e.g., Beta, Gaussian), for which Bayes update is simple to compute, it is possible to show that one can achieve an asymptotically optimal regret \citep{shipra, agrawal2017near}. 

The main idea behind the Batch Thompson Sampling (B-TS), outlined in Algorithm~\ref{alg:SI}, is as follows. 
For each arm $a\in [N]$,  B-TS keeps track of $\{k_a\}_{a\in[N]}$, the number of times the arm $a$ has been selected so far.  Initially, all $k_a$'s are set to 1. For each arm $a$ and at the beginning of the batch, necessarily $2^{l_a-1}\leq k_a< 2^{l_a}$ for some integer $l_a\geq 1$. Now consider a new batch that starts at time $t$. Within this batch, B-TS samples arms according to the prior distributions up to time $t-1$, namely $[D_a(t-1)]_{a\in[N]}$, and selects  the one with the highest value.  B-TS keeps selecting arms until the point that for one of the arms, say $a$, it reaches $k_a =2^{l_a}$. At this point, B-TS queries all the arms selected during this batch. Based on the received rewards, B-TS updates $\{D_a\}_{a\in[N]}$ and starts a new batch.

\begin{algorithm}[t!]
  \caption{~\bf{Batch Thompson Sampling}}
  \label{alg:SI} 
\begin{algorithmic}[1]
\State  \textbf{Initialize:} $k_a \leftarrow 0$ ($\forall a\in[N]$), $l_a\leftarrow 0$ ($\forall a\in[N]$), $\batch \leftarrow \emptyset$ 
\For {$t=1,2,\cdots T$}
\State   $\theta_a(t) \sim D_a(t)$ ($\forall a\in[N]$)
\toRemove{$Beta(S_i+1,F_i+1)$.}
\State  $a(t) := \argmax_{a\in[N]}\ \theta_a(t)$.
\State   $k_{a(t)} \leftarrow k_{a(t)} + 1$
\If {$k_{a(t)} < 2^{l_{a(t)}}$ }
\State $\batch \leftarrow \batch \cup \{a(t)\}$
\Else 
\State $l_{a(t)}=l_{a(t)}+1$
\State \text{Query($\batch$)} and receive rewards    
\toRemove{\State \text{Update $S,F$}}
\State \text{Update $D_a(t)$} ($\forall a\in \text{batch}$)
\State $\batch \leftarrow \emptyset$
\EndIf
\EndFor
\end{algorithmic}
\end{algorithm}

%% file: twoarm.tex



\paragraph{Regret Bounds with Beta Priors.}

For the ease of presentation, we first consider the Bernoulli  multi-armed bandit where $r_a\in\{0,1\}$ and $\mu_a =\Pr[r_a =1]$.  In this setting, we can instantiate TS with Beta priors as follows.  TS assumes an independent Beta-distributed prior, with parameters $(\alpha_a,\beta_a)$, over each $\mu_a$.  Due to the nice congugacy property of Beta distributions, it is very easy to update the posterior distribution, given the observations. In particular, the Bayes update can be performed as follows:
$$
(\alpha_a,\beta_a)=
\begin{cases}
(\alpha_a,\beta_a) & \text{if } a(t) \neq a,\\
(\alpha_a,\beta_a) + (r(t), 1-r(t)) & \text{if } a(t) = a.
\end{cases}
    $$
TS initially assumes $\alpha_a=\beta_a=1$ for all arms $a\in[N]$, which corresponds to the uniform distribution over $[0,1]$. The update rule of B-TS in Algorithm~\ref{alg:SI} is also very similar.  Let $B(t)$ be the last time $t'\leq t-1$ that B-TS carried out a batch. Moreover, for each arm $a$, let $S_a(t)$ be the number of instances arm $a$ was selected by time $t-1$ and $r_a=1$. Similarly, let $F_a(t)$ be the number of instances arm $a$ was selected by time $t-1$ and $r_a=0$. We also denote by $k_a(t) =S_a(t) + F_a(t)$ the total number of instances arm $a$ was selected by time $t-1$. Initially, B-TS starts with the uniform distribution over $[0,1]$, i.e., $D_a(1) = Beta(1,1)$ for all $a\in[N]$.  Inspired by the update rule of TS,  at any time  $t$, B-TS updates the distribution $D_a(t)$ by  $Beta(S_a(B(t))+1,F_a(B(t))+1)$. Note that during a batch when arms are being selected, the distributions $\{D_a\}_{a\in[N]}$ do not change. The updates only take place once the batch is carried out and the rewards are observed.



First we bound the number of batch queries as follows.

\begin{restatable}{theorem}{nq}\label{lem:nqq}
The total number of batches carried out by B-TS is at most $O(N\log T)$.
\end{restatable}
The  proof is given in Appendix~\ref{app:beta}.

\begin{figure*}[h] 
	\centering     
\subfloat[]{\includegraphics[height=50mm]{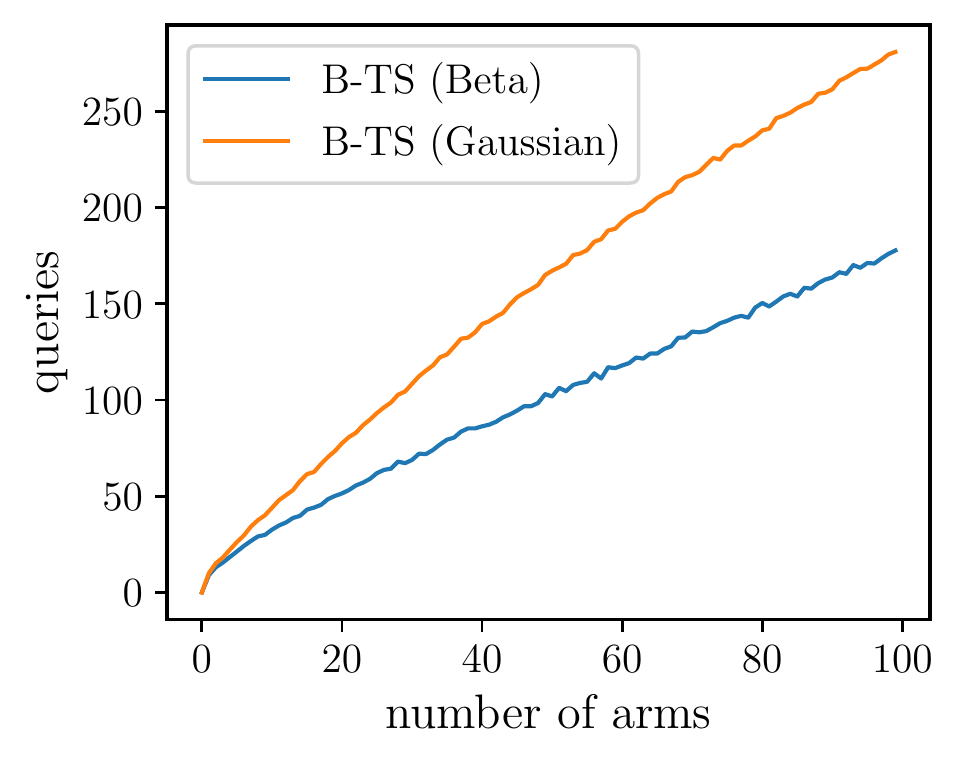}\label{single1}}
\subfloat[]{\includegraphics[height=50mm]{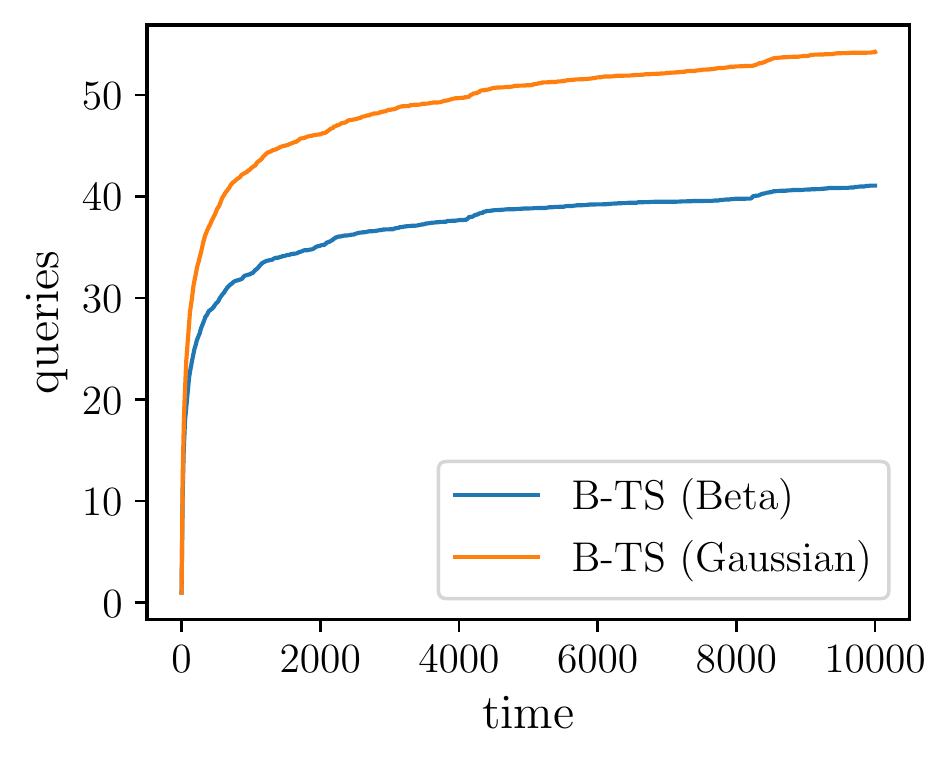}\label{single2}}
\caption{(a) and (b) show the number of batch queries versus the number of arms and the horizon, respectively.  We consider a synthetic Bernoulli setting where the horizon is set to  $T=10^3$ and  the number of arms vary from $N=1$ to $N=100$. We report the average regret over 100 experiments. As we clearly see in  Figure~\ref{single1},  the regret increases linearly in $N$ which rules out the possibility that the regret of B-TS may depend sub-linearly in $N$. For Figure~\ref{single2}, we also consider the Bernoulli setting and set $N=10$ and vary the horizon from $T=1$ to $T=10^4$. Again, as our theory suggests, the regret increases logarithmically in $T$.}
\label{fig:B-TS}
\end{figure*}

\begin{remark}
One might be tempted to show a sublinear dependency on $N$. However, simple empirical results show that the number of batches carried out by B-TS indeed scales logarithmically in $T$ but linearly in $N$. Please see figs \ref{single1} and \ref{single2} for more details.
\end{remark}

 What is more challenging is to show is that this simple batch strategy achieves the same asymptotic regret as a fully sequential one.

\begin{restatable}{theorem}{lns}\label{thm:ln}
Without loss of generality, let us assume that the first arm has the highest mean value, i.e.,  $\mu^*=\mu_1$. Then, the expected regret of B-TS, outlined in Algorithm~\ref{alg:SI}, with Beta priors can be bounded as follows
\[
\mathcal{R}(T) = (1+\epsilon) O\left( \sum_{a=2}^{N} \frac{\ln T}{d(\mu_a,\mu_1)}\Delta_a \right)+O\left(\frac{N}{\epsilon^2} \right),
\]
where
$d(\mu_a,\mu_1):=\mu_a\log \frac{\mu_a}{\mu_1} + (1-\mu_a)\log \frac{(1-\mu_a)}{1-\mu_1}$ and  $\Delta_a=\mu_1-\mu_a$. 
\end{restatable}
The complete proof is given in Appendix~\ref{app:beta}.

\begin{remark}
Even though we only provided the details for the Bernoulli setting, B-TS can be easily extended to general reward distributions supported over $[0,1]$. To do so, once a reward $r_t\in[0,1]$ is observed, we flip a coin with bias $r_t$ and update the Beta distribution according to the outcome of the coin. It is easy to see that Theorem~\ref{thm:ln} holds for this extension as well.
\end{remark}

\begin{remark}
\cite{gao2019batched} proved that for the $B$-batched $N$-armed bandit problem with time horizon $T$ it is necessary to have $B=\Omega(\log T / \log \log T)$ batches to achieve the problem-dependent asymptotic optimal  regret. This lower bound implies that B-TS use almost the minimum number of batches needed (i.e., $O(\log T)$ versus $\Omega({\log T}/{\log \log T})$) to achieve the optimal regret. 
\end{remark}
Now we present  the problem independent regret bound for B-TS.
\begin{restatable}{theorem}{root}
\label{thm:root}
Batch Thompson Sampling, outlined in Algorithm~\ref{alg:SI} and instantiated with Beta priors, achieves $\mathcal{R}(T)= O(\sqrt{NT \ln T})$ with $O(N\log T)$ batch queries. 
\end{restatable}

The proof is given in Appendix~\ref{app:betaIndep}.


\paragraph{Regret Bounds with Gaussian Priors.}
In order to obtain a better regret bound, we can instantiate TS with Gaussian distributions. To do so, let us define the empirical mean estimator for each arm $a$ as follows:
\[
 \hat{\mu}_a(t)= \frac{\sum_{\tau=1}^{t-1} r_{a(\tau)}\times \mathbb{I}(a(\tau)=a)}{k_a(t)+1 },
 \]
 where $k_a(t)$  denotes the number of instances that an arm $a\in[N]$ has been selected up to time $t-1$ and $\mathbb{I}(\cdot)$ is the indicator function.
 Then, by assuming that  the prior distribution of an arm $a$  is  $\mathcal{N}\left(\hat{\mu}_a(t),\frac{1}{k_a(t)+1}\right)$, and that the likelihood of $r_{a_t}$ given $\mu_a$ is $\mathcal{N}(\mu_a,1)$, the posterior will  also be a Gaussian distribution with parameters $\mathcal{N}\left(\hat{\mu}_a(t+1),\frac{1}{k_a(t+1)+1}\right)$. 
 
 In B-TS, we need to slightly change the way we estimate $ \hat{\mu}_a(t)$ as the algorithm has only access to the information received by the previous batches. Recall that $B(t)$ indicates the last time $t'\leq t-1$ that B-TS carried out a batch query. For each arm $a\in[N]$, we assume the prior distribution $D_a(t)\sim\mathcal{N}\left(\hat{\mu}_a(B(t)),\frac{1}{k_a(B(t))+1}\right)$. We also update the empirical mean estimator as follows: 
 \begin{align} \label{eq:update}
 \hat{\mu}_a(t)= \frac{\sum_{\tau=1}^{B(t)} r_{a(\tau)}\times \mathbb{I}(a(\tau)=a)}{k_a(B(t)+1)+1 }.
 \end{align}

 Note that  at any time $t$ during the $l$-th batch, i.e., $t\in(t_{l-1},t_{l}]$, the distribution $D_a(t)$ remains unchanged. Once the arms $\{a_t\}_{t\in (t_{l-1},t_{l}]}$  are carried out and the rewards $\{r_t\}_{t\in(t_{l-1},t_{l}]}$ are observed, $\hat{\mu}_a$ changes and the posterior is computed accordingly, namely, $D_a(t_{l}+1) \sim \mathcal{N}\left(\hat{\mu}_a(t_l+1),\frac{1}{k_a(t_l+1)+1}\right).$ As we instantiate B-TS with Gaussian priors, the regret bound slightly improves. 
\begin{restatable}{theorem}{Gaussian}
Batch Thompson Sampling, outlined in Algorithm~\ref{alg:SI} and instantiated with Gaussian priors, achieves $\Ex[\mathcal{R}(T)]= O(\sqrt{NT \ln N})$ with $O(N\log T)$ batch queries.
\end{restatable}
The proof is given in Appendix~\ref{app:gaussian}.

\section{Batch Minimax Optimal Thompson Sampling}
\begin{algorithm}[t!]
  \caption{~\bf{Batch Minimax Optimal Thompson Sampling (B-MOTS)}}
  \label{alg:MOTS} 
\begin{algorithmic}[1]
\State  \textbf{Initialize:} $k_a \leftarrow 0$ ($\forall a\in[N]$), $l_a\leftarrow 0$ ($\forall a\in[N]$), $\batch \leftarrow \emptyset$ 
\State \textbf{Initialize:} Play each arm $a$ once and initialize $D_a(t)$.
\For {$t=N+1,\cdots T$}
\State for all arms $a\in[N]$ sample
\begin{eqnarray*}
\tilde{\theta}_a(t)&\sim& \mathcal{N}(\hat{\mu}_a(B(t)), 1/(\rho k_a(B(t))))\\
\theta_a(t) &\sim& D_a(t)=\min\{\tilde{\theta}_a(t),\tau_a(t)\}
\end{eqnarray*}
\State $a(t):=\argmax_a \theta_a(t)$
\State   $k_{a(t)} \leftarrow k_{a(t)} + 1$ 
\If {$k_{a(t)} < 2^{l_{a(t)}}$ }
\State $\batch \leftarrow \batch \cup \{a(t)\}$
\Else 
\State $l_{a(t)}=l_{a(t)}+1$
\State Query($\batch$) \text{and observe rewards}
\State Update $D_a(t), \forall a\in [N]$
\State $\batch \leftarrow \emptyset$
\EndIf
\EndFor
\end{algorithmic}
\end{algorithm}

So far, we have considered the parallelization of the vanilla Thompson Sampling which does not achieve the optimal minimax regret. In this section, we introduce Batch Minimax Optimal Thompson Sampling (B-MOTS), that achieves the optimal minimax bound of  $O(\sqrt{NT})$, as well as the asymptotic optimal regret bound for Gaussian rewards. In contrast to the fully sequential MOTS developed by \citet{MOTS}, B-MOTS requires only  $O(N\log T)$ batches. The crucial difference between B-MOTS and B-TS is that instead of choosing Gaussian or Beta distributions, B-MOTS uses a clipped Gaussian distribution.

 To run B-MOTS, we need to slightly change the way  $D_a(t)$ is updated. First, to initialize $D_a(t)$, B-MOTS plays each arm once in the beginning and sets $k_a(N+1)$ to 1 and
$\hat{\mu}_a(N+1)$ to
the observed reward of each arm $a\in[N]$. To determine $D_a(t)$ for the subsequent batches, let  us first define a confidence range $(-\infty,\tau_a(t))$ for each arm $a\in [N]$ as follows:

\begin{equation}\label{eq:confidence}
    \tau_a(t) = \hat{\mu}_a(B(t)) + \sqrt{ \frac{\alpha}{k_a(B(t))} \log^+\left(\frac{T}{Nk_a(B(t))}\right)},
\end{equation}

where $\log^+(x) = \max\{0,\log(x)\}$ and the empirical mean  for each arm $a$ is estimated as 
 \begin{equation}\label{eq:muBMOTS}
 \hat{\mu}_a(t)= \frac{\sum_{\tau=1}^{B(t)} r_{a(\tau)}\times \mathbb{I}(a(\tau)=a)}{k_a(B(t)+1)}.
\end{equation}
Note that the estimators in \eqref{eq:muBMOTS} and \eqref{eq:update}
slightly differ due to the initialization step of B-MOTS. 

For each arm $a\in [n]$,  B-MOTS first samples  $\tilde{\theta}_a(t)$ from a Gaussian distribution with the following parameters 
$\tilde{\theta}_a(t)\sim \mathcal{N}(\hat{\mu}_a(B(t)), 1/(\rho k_a(B(t)))),$ where $\rho \in (1/2,1)$ is a tuning parameter. Then, the sample is clipped by the confidence range as follows:
\begin{equation}\label{eq:DBMOTS}
D_a(t)=\min\{\tilde{\theta}_a(t),\tau_a(t)\}. 
\end{equation}
The rest is exactly as in Alg~\ref{alg:SI} for B-TS. If you are interested in the details, you can find the outline of B-MOTS algorithm in Appendix~\ref{sec:B-MOTS-Proof}.

\paragraph{B-MOTS for SubGaussian Rewards.}

We state the regret bounds in the most general format, i.e.,  when the rewards follow a sub-Gaussian distribution. To remind ourselves, we say that a random variable $X$ is $\sigma$ sub-Gaussian if 
$\Ex[\exp(\lambda X - \lambda \Ex[X])] \le \exp( \sigma^2\lambda^2/2), \text{ for all } \lambda \in \mathbb{R}.$

The following theorem shows that B-MOTS is minimax optimal. 
\begin{restatable}{theorem}{mots}
\label{thm:minimax}
If the reward of each arm is 1-subgussian then the regret of B-MOTS  is bounded by $\mathcal{R}(T)=O(\sqrt{NT}+\sum_{a:\Delta_a>0} \Delta_a).$ Moreover, the number of batches is  bounded by $O(N\log T)$.
\end{restatable}

The proof is given in  Appendix~\ref{app:BMOTS}.



The next theorem presents the asymptotic regret bound of B-MOTS for sub-Gaussian rewards.

\begin{restatable}{theorem}{assymp}\label{thm:asymptotic}
Assume that the reward of each arm $a\in [N]$ is 1-subgaussian with mean $\mu_a$. For any fixed $\rho \in (1/2,1)$,
the regret of B-MOTS can be bounded as $ {\mathcal{R}(T)} = O \left(\log(T) \sum_{a:\Delta_a>0} \frac{1}{\rho\Delta_a} \right).$

\end{restatable}

The proof is given in  Appendix~\ref{app:assBMOTS}

The asymptotic regret rate of B-MOTS matches the existing lower bound $\log(T)\sum_{a:\Delta_a>0} 1/\Delta_a$ ~\citep{lai1985asymptotically} up to a multiplicative factor $1/\rho$. Therefore, similar to the analysis of the fully sequential setting \citep{MOTS}, B-MOTS reaches the exact lower bound at a cost of minimax optimality. In the next section, we show that at least in the Gaussian reward setting, minimax and asymptotic optimally cab be achieved simultaneously.


\paragraph{B-MOTS-J for Gaussian Rewards.}
In this part we present a batch version of Minimax Optimal Thompson Sampling for Gaussian rewards \citep{MOTS}, 
called B-MOTS-J, which achieves both minimax and asymptotic optimality when the reward distribution is Gaussian. The only difference between B-MOTS-J and B-MOTS is the way $\tilde{\theta}_a(t)$ are sampled. In particular, B-MOTS-J samples $\tilde{\theta}_a(t)$ according to $\mathcal{J}(\mu, \sigma^2)$ (instead of a Gaussian distribution), where the PDF is defined as 
\[
\Phi_{\mathcal{J}}(x) = \frac{1}{2\sigma^2} |x-\mu|\exp \left [-\frac{1}{2}\left(\frac{x-\mu}{\sigma}\right)^2 \right].
\]
Note that when $x$ is restricted to  $x\ge 0$, then $\mathcal{J}$ becomes a Rayleigh distribution. More precisely, to sample $\tilde{\theta}_a(t)$, we set the parameters of $\mathcal{J}$ as follows:
 $\tilde{\theta}_a(t)\sim \mathcal{J}\left(\hat{\mu}_a (B(t)),\frac{1}{k_a(B(t))}\right),$
where $\hat{\mu}_a (t))$ is estimated according to \eqref{eq:muBMOTS}. The rest of the algorithm is run exactly like B-MOTS.


\begin{restatable}{theorem}{motsj}
\label{thm:MOTS-J}
Assume that the reward of each arm $a$ is sampled from a Gaussian distribution $\mathcal{N}(\mu_a,1)$ and $\alpha>2$. 
Then, the regret of B-MOTS-J can be bounded  as follows:
$$
\mathcal{R}(T) = O(\sqrt{KT}+\sum_{a=2}^k \Delta_a), \quad
 \lim_{T \rightarrow \infty} \frac{\mathcal{R}(T)}{\log(T)} = \sum_{a:\Delta_a>0} \frac{2}{\Delta_a}.
$$

\end{restatable}
The proof is given in  Appendix~\ref{app:BMOTSJ}.


\section{Batch Thompson Sampling  for Contextual Bandits}
\begin{algorithm}[t!]
  \caption{~\bf{Batch TS for Contextual Bandits}}
  \label{alg:CB} 
\begin{algorithmic}[1]
\State  \textbf{Initialize:} $k_a \leftarrow 0$ ($\forall a\in[N]$), $l_a\leftarrow 0$ ($\forall a\in[N]$), $\batch \leftarrow \emptyset$, $B=I_d$, $\hat{\mu}=0_d$ 
\For {$t=1,2,\cdots T$}
\State  $\tilde{\mu}(t)\sim \mathcal{N}(\hat{\mu}(B(t)),v^2{\cal{B}}(B(t))^{-1})$
\State $a(t)=\argmax_a b_a(t)^T \tilde{\mu}(t)$
\State   $k_{a(t)} \leftarrow k_{a(t)} + 1$

\If {$k_{a(t)} < 2^{l_{a(t)}}$ }
\State $\batch \leftarrow \batch \cup \{a(t)\}$
\Else 
\State $l_{a(t)}=l_{a(t)}+1$
\State Query(batch) \text{and receive rewards }
\State \text{Update 
$\hat{\mu}$} 
\State $\batch \leftarrow \emptyset$
\EndIf
\EndFor
\end{algorithmic}
\end{algorithm}
In this section, we propose Batch Thompson Sampling for Contextual Bandits (B-TS-C), outlined in Algorithm~\ref{alg:CB}. As in the fully sequential TS, proposed by \citet{contextual}, we assume Gaussian priors and Gaussian likelihood functions. However, we should highlight that the analysis of B-TS-C and the corresponding regret bound hold irrespective of whether or not the  reward distribution matches the Gaussian priors and Gaussian likelihood functions (similar to the multi-armed bandit setting discussed in Section~\ref{sec:alg}). More formally, given a context $b_a(t)$, and parameter $\mu$, we assume that the likelihood of the reward $r_a(t)$ is given by $\mathcal{N}(b_a(t)^T\mu, v^2)$, where $v=\sigma\sqrt{9d \ln(T/\delta)}$ and $\delta \in (0,1)$\footnote{If the horizon $T$ is unknown, we can use $v_t=\sigma\sqrt{9d \ln(t/\delta)}$.}. Let us define the matrix ${\cal{B}} (t) $ as follows
\begin{align*}
{\cal{B}}(t)= I_d+ \sum_{\tau=1}^{t-1}b_{a(\tau)}(\tau)b_{a(\tau)}(\tau)^T.
\end{align*}
Note that the matrix $\mathcal{B}(t)$ depends on all the contexts observed up to time $t-1$. We consider the prior $\mathcal{N}(\hat{\mu}(B(t)),v^2{\cal {B}}(B(t))^{-1})$ for $\mu$ and update the the empirical mean estimator as follows: 
\begin{align}
    \hat{\mu}(t)&={\cal{B}}(B(t))^{-1} \left(\sum_{\tau=1}^{B(t)} b_{a(\tau)}(\tau) \times r_{a(\tau)}(\tau) \right).
\end{align}
Note that in order to estimate $\hat{\mu}(t)$, we only consider the rewards received up to time $B(t)$, namely, the rewards of arms pulled in the previous batches.  At each time step $t$, B-TS-C generates  a sample $\tilde{\mu}(t)$ from
$\mathcal{N}(\hat{\mu}(B(t)),v^2{\cal{B}}(B(t))^{-1})$ 
and plays the arm $a$ that maximizes  $b_a(t)^T\tilde{\mu}(t)$. 
The posterior distribution for $\mu$  at time $t+1$ will be $\mathcal{N}(\hat{\mu}(B(t+1)), v^2{\cal B}(B(t+1))^{-1})$.

\begin{restatable}{theorem}{contextual} \label{thm:contextual}
The B-TS-C algorithm (Algorithm~\ref{alg:CB})
achieves the total regret 
of 
$$\mathcal{R}(T)=O\left(d^{3/2}\sqrt{T}(\ln (T)+\sqrt{\ln (T) \ln (1/\delta)})\right)$$
with probability $1-\delta$. 
Moreover, B-TS-C carries out $O(N\log T)$ batch queries.
\end{restatable}
The proof is given in  Appendix~\ref{app:CB}.


\vspace{-0.3cm}
\section{Experimental Results}
\vspace{-0.3cm}

\begin{figure*}[t] 
	\centering     
\subfloat[]{\includegraphics[height=35mm]{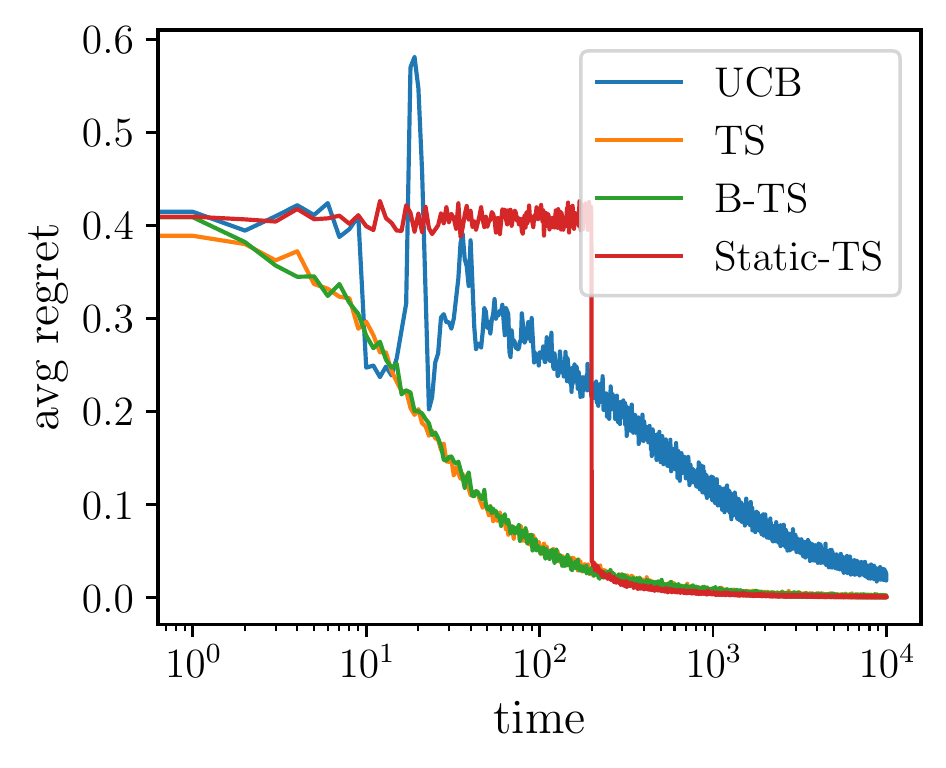}\label{twitterUtil}}
\subfloat[]{\includegraphics[height=35mm]{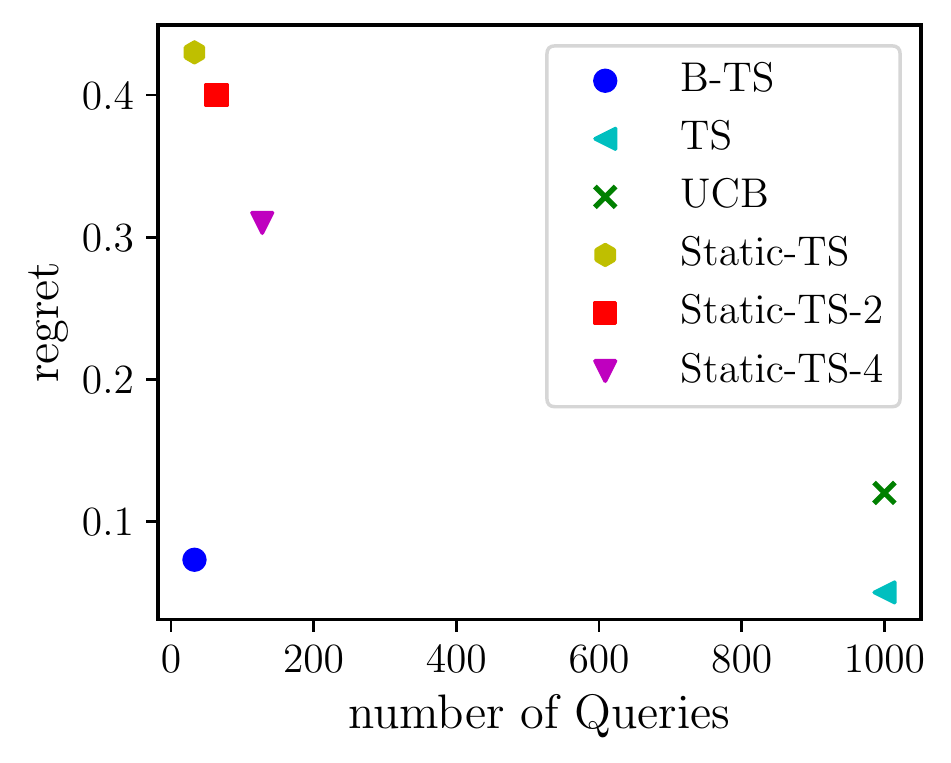}\label{varyEps2}}	
\subfloat[]{\includegraphics[height=38mm]{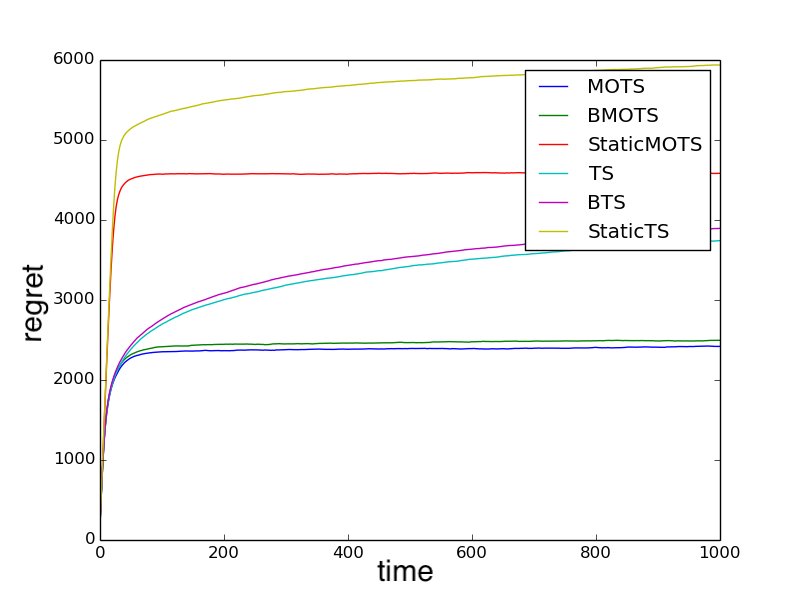}\label{single3}}\\
\subfloat[]{\includegraphics[height=38mm]{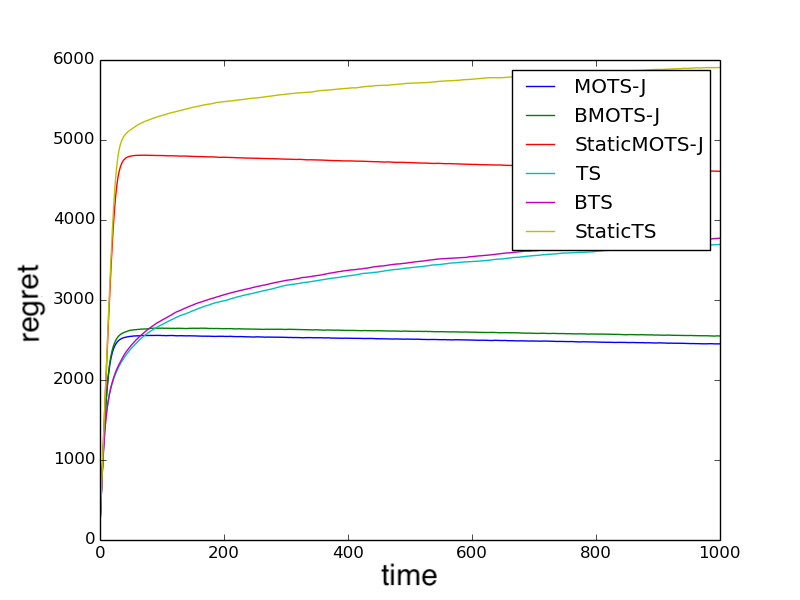}\label{single4}}
\subfloat[]{\includegraphics[height=36mm]{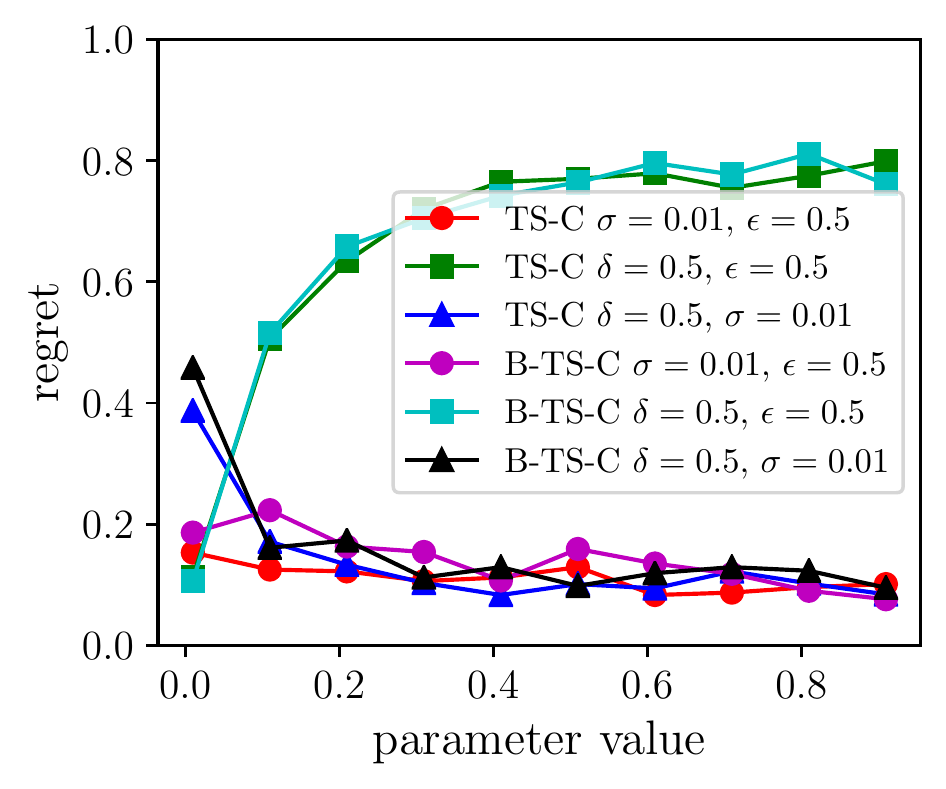}\label{twitterAdapt}}
\subfloat[]{\includegraphics[height=36mm]{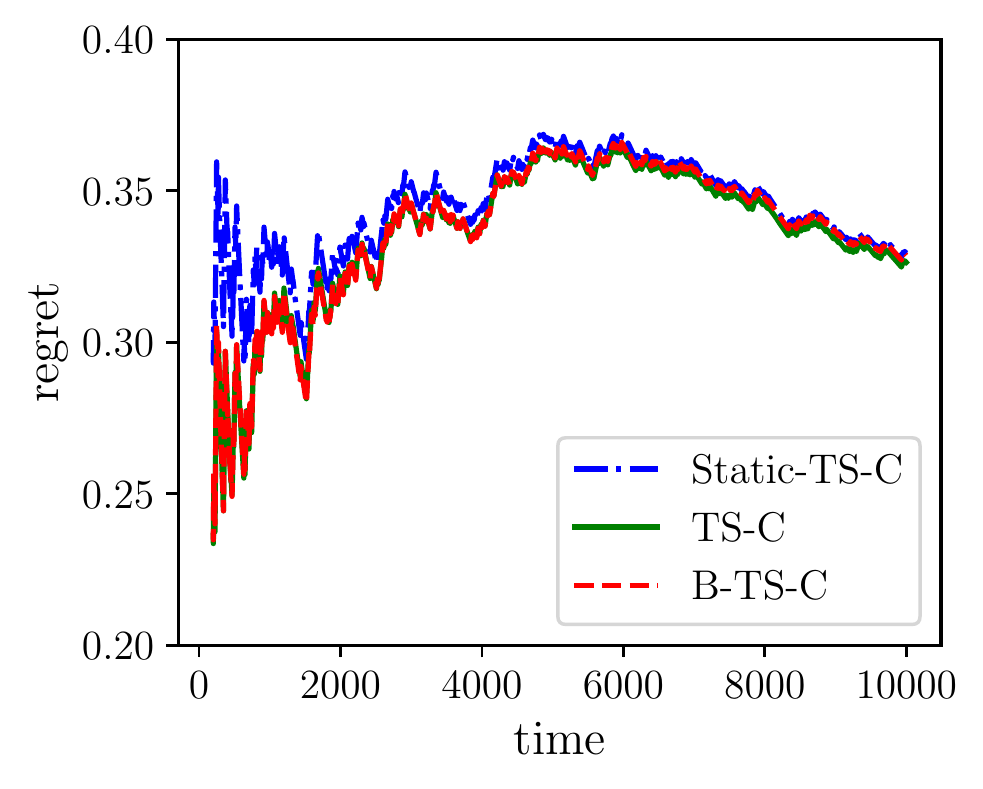}\label{varyEps1}}
\caption{ (a) and (b) compare the regret of UCB against TS and its batch variants. (c) and (d) compare the batch variants of TS and MOTS. (e) shows the sensitivity of TS-C and its batch variants to the tuning parameters. (f) shows the performance of TS-C and its batch variants on real data.\vspace{-0.5cm}}
\label{testGraphs}
\end{figure*}

In this section, we compare the performance of our proposed batch Thompson Sampling policies (e.g., B-TS,B-MOTS, B-MOT-J and B-TS-C) with their fully sequential counterparts. We also include several baselines such as  UCB  and Thompson Sampling with static batch design (Static-TS). In particular, for Static-TS, Static-TS2, and  Static-TS4, we set the total number of batches to that of B-TS, twice of B-TS, and four times of B-TS, respectively.  However,   in the static batch design, we use  equal sized batches.

\paragraph{Batch Thompson Sampling.}
In Figure~\ref{twitterUtil}, we compare the performance of UCB against TS and its batch variants in a synthetic Bernoulli setting. We vary $T$ from 1 to $10^4$ and set $N=10$. We run all the experiments 1000 times. Figure~\ref{twitterUtil} compares the average regret, i.e., $\mathcal{R}(T)/T$, versus the horizon $T$.  As expected, TS outperforms UCB. Moreover, TS and B-TS follow the same trajectory and have practically the same regret. Note that for the static variant of TS, namely, Static-TS, we see that in the first few hundred iterations, its performance is even worst than UCB and then it catches with TS. Figure~\ref{varyEps2} more clearly shows the trade-off between the regret obtained by different baselines versus the number of batch queries (bottom-left is the desirable location). In this figure, we set $T=10^3$. We see that the lowest regret is achieved by TS and B-TS but TS carries out many more queries. Also, we should highlight that while the static versions make fewer queries than TS, they do not achieve a similar regret. Notably, even Static-TS-4, that carries out 4 times more queries than B-TS, has a much higher regret. This shows the importance of a dynamic batch design.
\paragraph{Batch Minimax Optimal Thompson Sampling.}
In Figures~\ref{single3} and \ref{single4} we compare the regret of MOTS \cite{MOTS} and its batch versions. The synthetic setting 
is similar to~\cite{MOTS}. We set $N=50$, $T=10^6$ and the total number of runs to 2000. The reward of each arm is sampled from an independent Gaussian distribution. More precisely, the optimal arm has the expected reward and variance 1 while the other $N-1$ arms have the expected reward $1-\epsilon$ and variance 1 (we set $\epsilon=0.2$). For MOTS, we set $\rho=0.9999$ and $\alpha=2$ as suggested by \citet{MOTS}. As we can see in Figure~\ref{single3}, the batch variants of TS and MOTS achieve practically a similar regret. Also, as our theory suggests, B-MOTS (along with MOTS) have the lowest regret while B-MOTS drastically reduces the number of batches w.r.t MOTS. Moreover, the static batch designs, namely Static-TS and Static-MOTS, show the highest regret while carrying out the same number of batch queries as B-TS and B-MOTS. Therefore, the dynamic batch design of B-TS and B-MOTS seems crucial for obtaining good performances. A similar trend is shown in Figure~\ref{single4} where we run MOTS-J (with $\alpha=2$) and its batch variants. Again, B-MOTS-J and MOTS-J are practically indistinguishable  while achieving the lowest regret.


\paragraph{Contextual Bandit.}
For the contextual bandit, we perform a synthetic and a real-data experiment. Figure~\ref{twitterAdapt} shows the performance of the sequential Thompson Sampling, namely, TS-C, and the batch variants, namely, B-TS-C, as we change different parameters $\epsilon,\delta$ and $\sigma$ from 0 to 1. Here, the context dimension is 5 and we set the horizon to $T=10^4$. We run all the experiments 1000 times. As we see in Figure~\ref{twitterAdapt}, TS-C and B-TS-C follow practically the same curves. 

For the  experiment on real data, we use the MovieLens data set where the dimension of the context is 20, the horizon is $T=10^5$, and we run each experiment 100 times. For the parameters, we set $\delta=0.61, \sigma=0.01$, and  $\epsilon=0.71$ as suggested by \citet{beygelzimer2011contextual}. We see that Static-TS-C performs a bit worst than  TS-C and B-TS-C, again suggesting that it is crucial to use dynamic batch sizes.

%% file: conclusion.tex
\vspace{-0.4cm}
\section{Conclusion}
\label{sec:conclusion}
\vspace{-0.3cm}
In this paper, we revisited the classic Thompson Sampling procedure and developed the first Batch variants for the stochastic multi-armed bandit and linear contextual bandit. We proved that our proposed batch policies achieve similar regret bounds (up to constant factors) but with significantly fewer number of interactions with the environment. We have also  demonstrated experimentally that our batch policies achieve practically the same regret on both synthetic and real data.

%% file: appendix.tex
\section{Batch Thompson Sampling for Multi-armed Bandit}

In this section, we follow the notations used in \cite{shipra,agrawal2017near} and adapt them to the batch setting.

\subsection{Notations and Definitions}
\label{app:BTSNOT}

\begin{definition}
For a Binomial distribution with parameters $\alpha$ and $\beta$,  we refer to its CDF as $F_{n,p}^{B}(.)$, and pdf as $f_{n,p}^{B}(.)$. 
We furthermore denote by $F_{\alpha,\beta}^{beta}(.)$ the CDF of Beta distribution.
It is easy to show that 
for all $\alpha,\beta>0$,
\[
F_{\alpha, \beta}^{beta}(y)=1-F_{\alpha+\beta-1,y}^B(\alpha-1) \ .
\]

\end{definition}

\begin{definition} [History/filtration ${\cal F}_t$]
For time steps $t=1,\cdots, T$ define the history of the arms that have been played upto time $t$  as
\[
{\cal F}_t=\{a(\tau),r_{a(t)}(\tau), \tau \leq t \} \ .
\]
\end{definition}

\begin{definition}\label{def:tauA}
For a given arm $a$, we  denote by $\tau_j$ the time step in which $a$
has been \textit{queried} for the $j$-th time. We let $\tau_0=0$.
Note that $\tau_T\ge T$.
\end{definition}

\begin{definition}
Denote by $\theta_a(t)$ the sample for arm $a$ at time $t$ from the posterior distribution at time $B(t)$, 
namely $Beta(S_a(B(t))+1,k_a(B(t))-S_a(B(t))+1)$.
\end{definition}

\begin{definition}\label{def:thresholds} Without loss of generality, we assume that $a=1$ is the optimal arm.
For a non-optimal arm $a\ne 1$, 
we have two thresholds $x_a, y_a$ depending on the type of upper bounds we are proving (i.e., problem dependent or independent) such that
$\mu_a <x_a < y_a <\mu_1$. 
\end{definition}

\begin{definition} \label{def:da}
We  denote by $\Delta'_a:=\mu_1-y_a$
and $D_a:=y_a \ln\frac{y_a}{\mu_1}+(1-y_a)\ln \frac{1-y_a}{1-\mu_1}$.
Also define $d(\mu_a,\mu_1):=\mu\log \frac{\mu_a}{\mu_1}+(1-\mu_a)\log \frac{1-\mu_a}{1-\mu_1}$.

\end{definition}

\begin{definition}\label{def:EA}
For a non-optimal arm $a$ (i.e., $a\ne 1$), we use  $E_a^{\mu}(t)$ for the event $\{\hat{\mu}_a(B(t))\le x_a\}$ and we use $E_a^{\theta}(t)$ for the event  $\{\theta_a(t)\le y_a\}$.
\end{definition}

\begin{definition} \label{def:p}
The
(conditional) probability that for a non optimal arm $a$, the generated sample for the optimal arm $a=1$  at time  $t$ exceeds the threshold $y_a$ is defined as
\[
p_{a,t}:=\Pr(\theta_1(t)>y_a | \mathcal{F}_{B(t)}) \ .
\]
\end{definition}
Here is our first lemma regarding the relationship between batch bandit and sequential bandit. 
\begin{lemma}\label{lem:halfsize}
For any arm $a$, we have $k_a(B(t))\ge \frac{1}{2}k_a(t)$.
\end{lemma}
\begin{proof}
The reason is that if $k_a(B(t))< \frac{1}{2}k_a(t)$ then B-TS (Algorithm~\ref{alg:SI})  should have queried a batch after time $B(t)$ which is a contradiction.
\end{proof}


\subsection{Problem-dependent Regret Bound with Beta Priors}
\label{app:beta}

\nq*

\begin{proof}
Every time we query a batch, 
there is one arm $a$, for which
$k_a=2^{\ell_a}$.
In order to count the total number of batches, we assign each time step $t$ to a batch $B$. Note that the assigned batch for $t$ is not necessarily the batch that $a(t)$ will be added to. 
Suppose $k_a=2^{\ell_a}$, and the algorithm queries a batch $B$, we assign time steps 
in which arm $a$ was queried for the $2^{\ell_a-1}+1,\cdots, 2^{\ell_a}$-th times to the batch $B$ (although some of the elements might have been queried in the previous batches). Let's denote this set by $T_a(B)$. Then for each arm $a$, the total number of batches corresponding to arm $a$ is at most $O(\log T)$ (since the last time step arm $a$ is being played is at most $T$).
Therefore, we can upper bound the total number of batches by $O(N\log T)$ batches.
\end{proof}

First, note that in the batch algorithm B-TS (Algorithm~\ref{alg:SI}),  we define $\theta_a(t)$ based on $\mathcal{F}_{B(t)}$. As a result of  these modifications the following lemma is immediate. It is a batch variation of \cite[Lemma 2.8]{agrawal2017near}.

{
\begin{lemma} \label{lem:prarm}
For all $t$,all suboptimal arm $a\ne 1$, and all instantiation $\mathcal{F}_{B(t)}$  we have
\[
\Pr(a(t)=a,E_{a}^{\mu}(t), E_{a}^{\theta}(t)| \mathcal{F}_{B(t)}) \le \frac{1-p_{a,t}}{p_{a,t}} \Pr(a(t)=1, E_{a}^{\mu}(t),E_{a}^{\theta}(t)|\mathcal{F}_{B(t)}) \ .
\]
\end{lemma}
\begin{proof}
$E_a^{\mu}(t)$ is determined by $\mathcal{F}_{B(t)}$. Therefore it is enough to 
show that for any instantiation $\mathcal{F}_{B(t)}$  
\[
\Pr(a(t)=a|E_{a}^{\theta}(t),\mathcal{F}_{B(t)}) \le \frac{1-p_{a,t}}{p_{a,t}} \Pr(a(t)=1|E_{a}^{\theta}(t),\mathcal{F}_{B(t)}) \ .
\]
Now given $E_a^{\theta}(t)$, we have $a(t)=a$ only if $\theta_j(t) \le y_a, \forall j$. 
Therefore, for $a\ne 1$ and any instantiation $\mathcal{F}_{B(t)}$ we have 
\begin{align*}
\Pr(a(t)=a|E_{a}^{\theta}(t),\mathcal{F}_{B(t)}) &\le \Pr(\theta_j(t)\le y_a, \forall j | E_a^{\theta}(t), \mathcal{F}_{B(t)})\\
&=\Pr(\theta_1(t)\le y_a | \mathcal{F}_{B(t)}).\Pr(\theta_j(t)\le y_a, \forall j\ne 1 | E_a^{\theta}(t),\mathcal{F}_{B(t)})\\
&=(1-p_{a,t}).\Pr(\theta_j(t)\le y_a,\forall j\ne 1 | E_{a}^{\theta}(t),\mathcal{F}_{B(t)}) \ .
\end{align*}
In the first equality given $\mathcal{F}_{B(t)}$, the random variable $\theta_1(t)$ is independent of all other $\theta_j(t)$ and $E_{a}^{\theta}(t)$. The argument for $a=1$ is similar.
\end{proof}
}

Now we prove the main lemma which provides a problem-dependent upper bound on the regret.

\lns*

\begin{proof}
The proof closely follows \cite[Theorem 1.1]{agrawal2017near} and is adapted to the batch setting. 
For a non optimal arm $a\ne 1$, we decompose the expected number of plays of arm $a$ as follows
\begin{align}\label{eq:ek}
\Ex\left[k_a(t)\right]&=\sum_{t=1}^T \Pr(a(t)=a)  \nonumber\\
&=\sum_{t=1}^{T} \Pr(a(t)=a,E_a^{\mu}(t),E_a^{\theta}(t))+ \sum_{t=1}^{T} \Pr(a(t)=a, E_a^{\mu}(t),\overline{E_a^{\theta}(t)})+
\sum_{t=1}^{T} \Pr(a(t)=a,\overline{E_a^{\mu}(t)})\ .
\end{align}

The first term can be bounded by lemma~\ref{lem:prarm} as follows:
\begin{align*}
\sum_{t=1}^{T} \Pr(a(t)=a,E_a^{\mu}(t),E_{a}^{\theta}(t))
&\le  \sum_{t=1}^{T} \Ex\left[\Pr(a(t)=a,E_a^{\mu}(t),E_a^{\theta}(t)|\mathcal{F}_{B(t)})\right] \\
& \le \sum_{t=1}^{T} \Ex\left[\frac{(1-p_{a,t})}{p_{a,t}}\Pr(a(t)=1,E_a^{\theta}(t),E^{\mu}_a(t))|\mathcal{F}_{B(t)}\right] \\
&= \sum_{t=1}^{T} \Ex\left[\Ex\left[\frac{1-p_{a,t}}{p_{a,t}}I(a(t)=1,E_a^{\theta}(t),E_a^{\mu}(t))|\mathcal{F}_{B(t)}\right]\right]\\
&\le \sum_{t=1}^{T} \Ex\left[\frac{1-p_{a,t}}{p_{a,t}} I(a(t)=1,E_a^{\theta}(t),E_{a}^{\mu}(t))\right] \ . \\
\end{align*}
Note that as before, given $\mathcal{F}_{B(t)}$, the probability $p_{a,t}$ is fixed which implies the second inequality. The difference between this argument and that of The proof closely follows \cite[Theorem 1.1]{agrawal2017near} is that conditioning is until the last time the B-TS algorithm has queried a batch, i.e., $B(t)$. 
Note that $p_{a,t}=\Pr(\{\theta_1(t)>y_a|\mathcal{F}_{B(t)}\})$ changes only after a batch queries 
the optimal arm.
Hence as before $p_{a,t}$ remains the same at all time steps $t\in \{\tau_k+1,\cdots, \tau_{k+1}\}$ (refer to Definition~\ref{def:tauA}).  Thus we can get the following decomposition 

\begin{align}\label{eq:pi}
\sum_{t=1}^T \Ex\left[\frac{1-p_{a,t}}{p_{a,t}}\mathbb{I}(a(t)=1, E_a^{\theta}(t),E_a^{\mu}(t))\right] &\le \sum_{k=0}^{T-1}\Ex\left[\frac{(1-p_{a,\tau_k+1})}{p_{a,\tau_k}+1}\sum_{t=\tau_k+1}^{\tau_{k+1}} \mathbb{I}(a(t)=1,E_a^{\theta}(t),E_a^{\mu}(t))\right]\nonumber\\
&\le \sum_{k=0}^{T-1} \Ex\left[\frac{1-p_{a,\tau_{k}+1}}{p_{a,\tau_k+1}}\right] \ .
\end{align}

Now for the term $\Ex\left[\frac{1}{p_{a,\tau_k+1}}\right]$, since $k_a(B(t))\ge 1/2k_a(t)$ (Lemma~\ref{lem:halfsize}), we can get a modification of the bound provided in \citet[Lemma 2.9]{agrawal2017near}, as follows. 

\begin{lemma}
Let $\tau_k$ be  the time step that optimal arm 1 has been played for the $k$-th time,
Then for non optimal arm $a\ne 1$ we have,
\[
\Ex\left[\frac{1}{p_{a,\tau_k}+1}-1\right]=\begin{cases}
      \frac{3}{\Delta'_a}, & \text{for }  k<\frac{16}{\Delta'_a}, \\
      \Theta\left(\exp(-\Delta_{a}^{'2}k/4)+\frac{\exp(-D_a k/2)}{(k/2+1)\Delta_{a}^{\prime 2}} +\frac{1}{\exp(\Delta_a^{\prime 2}k/16)-1}\right), 
      & \text{otherwise.}
    \end{cases}
\]
\end{lemma}

Similar to~\citet[Lemma 2.10]{agrawal2017near}, we  obtain the following lemma. 

\begin{lemma}\label{lem:Emutheta}
For a non optimal arm $a\ne 1$, we have
\[
\sum_{t=1}^{T} \Pr(a(t)=a,E_a^{\mu}(t),E_a^{\theta}(t)) \le \frac{48}{\Delta_a'^2} + \sum_{j>16/\Delta'_a} \Theta\left(e^{-\Delta_a'^2j/4}+\frac{2}{(j+1)\Delta_a'^2}\right) e^{-D_aj/2} + \frac{1}{e^{\Delta_a'^2j/8}-1}.
\]
\end{lemma}
Now by substituting the above lemma into equation~\ref{eq:pi},
we can upper bound other terms in equation~\eqref{eq:ek} to prove the following lemma.

\begin{lemma}\label{lem:Enotmu}
For a non optimal arm $a\ne 1$, we have
\[
\sum_{t=1}^{T} \Pr(a(t)=a,\overline{E_a^{\mu}(t)}) \le \frac{2}{d(x_a,\mu_a)}+1 \ .
\]
\end{lemma}
\begin{proof}
Let $\tau_k$ be the $k$-th play of arm $a$. 
The LHS can be upper bounded by  $\sum_{k=0}^{T-1} \Pr(\overline{E_a^{\mu}(\tau_{k+1})}) $.
Note that $\hat{\mu}_a$ will be updated when the algorithm queries a batch.
Using Chernoff-Hoeffding bound $$\Pr(\hat{\mu}_a(B(\tau_{k+1})) > x_a) \le e^{-\frac{1}{2}kd(x_a,\mu_a)},$$
where $x_a$ is defined in Definition~\ref{def:thresholds}.
Note that at time $B(\tau_{k+1})$, arm $a$ has been played at least $k/2$ times.
Thus,
\[
\sum_{t=1}^{T} \Pr(\overline{E_a^{\mu}(\tau_{k+1})}) = \sum_{k=0}^{T-1} \Pr(\hat{\mu}_a(B(\tau_{k+1}))>x_a)  \le 1+\sum_{k=1}^{T-1}\exp(-\frac{1}{2}kd(x_a,\mu_a)) \le 1+ \frac{2}{d(x_a,\mu_a)}.
\]
\end{proof}
The statement of the following lemma is similar to \cite[Lemma 2.12]{agrawal2017near}. However, we prove it for the batch policy. 
\begin{lemma} \label{lem:UPL}
For a non optimal arm $a\ne 1$, we have
\[
\sum_{t=1}^{T} \Pr(a(t)=a, \overline{E_a^{\theta}(t)},E_a^{\mu}(t)) \le L_a(t)+1,
\]
where $L_a(t)=\frac{\ln T}{d(x_a,y_a)}$. 
\end{lemma}
\begin{proof}
We can consider two cases when $k_a(B(t))$ is large (greater than $L_a(t)$) or small (less than $L_a(t)$). This way, we have
\begin{align}
\label{eq:decompose}
\sum_{t=1}^{T} \Pr(a(t)=a, \overline{E_a^{\theta}(t)},E_a^{\mu}(t)) = \sum_{t=1}^{T} \Pr(a(t)=a, k_a(B(t))\le L_a(t),\overline{E_a^{\theta}(t)},E_a^{\mu}(t) ) \nonumber \\
+ \sum_{t=1}^{T} \Pr(a(t)=a, k_a(B(t))>L_a(t), \overline{E_a^{\theta}(t)},E_a^{\mu}(t)).
\end{align}
Same as before the first term is bounded by 
$\Ex\left[\sum_{t=1}^T \mathbb{I}(a(t)=a,k_a(B(t))\le L_a(t))\right]$ which is bounded by
$L_a(t)$. Again we bound the second term by 1.
The main idea is to show that for large enough $k_a(B(t))$, and given $E_a^{\mu}(t)$ is true,  the probability of $E_a^{\theta}(t)$ being false is small. 
We can write
\begin{align*}
\sum_{t=1}^{T} \Pr(a(t)=a, k_a(B(t))>L_a(t), \overline{E_a^{\theta}(t)},E_a^{\mu}(t) ) 
&= \Ex\left[\sum_{t=1}^T \mathbb{I}(k_a(t)>L_a(t),E_a^{\mu}(t)) \Pr(a(t)=a,\overline{E_a^{\theta}}(t)|\mathcal{F}_{B(t)})\right] \\
&\le \Ex\left[\sum_{t=1}^{T} \mathbb{I}(k_a(t)>L_a(t),\hat{\mu_a(B(t))\le x_a}) \Pr(\theta_a(t)>y_a|\mathcal{F}_{B(t)})\right].
\end{align*}
Note that $\mathcal{F}_{B(t)}$ determines both $k_a(B(t))$ and $E_a^{\mu}(t)$.
Now, $\theta_a(t)$ is distributed according to  $$\theta_a(t) \sim Beta(\hat{\mu}_a(B(t))k_a(B(t)+1,(1-\hat{\mu}_a(B(t))k_a(B(t)))).$$
Given $E_a^{\mu}(t)$, it is stochastically dominated by
$Beta(x_a k_a(B(t))+1,(1-x_a)k_a(B(t)))$.
Now, if $\mathcal{F}_{B(t)}$ contains the events $E_a^{\mu}(t)$ and $\{k_a(B(t)) > L_a(t)\}$, we have
\[
\Pr(\theta_a(t)>y_a | \mathcal{F}_{B(t)}) \le 1- F_{ x_a k_a(B(t))+1, (1-x_a)k_a(B(t))  }^{beta} (y_a) \ .
\]
Using 
the Chernouf-Hoefding inequality, we can show that
the RHS of the above inequality is at most
\begin{align*}
1- F_{ x_a k_a(B(t))+1, (1-x_a)k_a(B(t))  }^{beta} (y_a)&=F^{B}_{k_a(B(t))+1,y_a}(x_a(k_a(t)+1)) \\
&\le \exp(-(k_a(B(t))+1)d(x_a,y_a))\\
&\le \exp(-(L_a(t))d(x_a,y_a)) \\
& \le 1/T \ .
\end{align*}

Summing over $t$  yields the upper bound 1 for the second term in~\ref{eq:decompose}.
\end{proof}

The rest of the proof is by combining the above lemmas and by setting the right value for $x_a$ and $y_a$ as discussed in~\cite{agrawal2017near}.
In particular, by combining Lemma~\ref{lem:Emutheta},~\ref{lem:Enotmu}, and~\ref{lem:UPL}
we have
\[
\Ex\left[k_a(t)\right] \le \frac{48}{\Delta_a'^2} + \sum_{j>16/\Delta'_a} \Theta(e^{-\Delta_a'^2j/4}+\frac{2}{(j+1)\Delta_a'^2}) e^{-(D_{a})j/2} + \frac{1}{e^{\Delta_a'^2j/8}-1}) + L_a(t)+1+\frac{1}{d(x_a,\mu_a)}+1 \ .
\]
Now we should set the  right value to parameters $x_a,y_a$.
For $0\le \epsilon <1$, set $x_a \in (\mu_a,\mu_1)$ such that $d(x_a,\mu_1)=d(\mu_a,\mu_1)/(1+\epsilon)$ and set $y_a\in (x_a,\mu_1)$ such that $d(x_a,y_a)=d(x_a,\mu_1)/(1+\epsilon)=d(\mu_a,\mu_1)/(1+\epsilon)^2$. For these values, the regret bound easily follows. 
We will use different values for problem independent case in the next section.

\end{proof}

\subsection{Problem-independent Regret Bound with Beta Priors}
\label{app:betaIndep}

Now we prove  the problem independent regret bound.

\root*

\begin{proof}
The proof follows \cite[Theorem 1.2]{agrawal2017near} and adapted to the batch setting. 
For each sub-optimal arm $a\ne 1$, in the analysis of the algorithm we use two thresholds $x_a$ and $y_a$ such that 
$\mu_a < x_a < y_a < \mu_1$.
These parameters respectively control the events that the estimate $\hat{\mu}_a$ and sample $\theta_a$ are not too far away from the mean of arm $a$, namely, $\mu_a$. To remind the notation in  Definition~\ref{def:EA},
$E_a^{\mu}(t)$ represents the event $\{ \hat{\mu}_a(B(t))\le x_a \}$ and  $E_a^{\theta}(t)$ represents the event $\{ \theta_a(t)\le y_a \}$. The probability of playing each arm will be upper bounded based on whether or not   the above events are satisfied.

Furthermore, the threshold $y_a$ is also used in the definition of $p_{a,t}$ (see Definition~\ref{def:p}) 
and Lemma~\ref{lem:prarm}  to bound 
 the probability of playing any suboptimal arm $a\ne 1$ at the current step $t$ by a linear function of $p_{a,t}$.
Additionally, in Lemma~\ref{lem:Enotmu} we show an upper bound for the probability of selecting arm $a$ in terms of $x_a$ and $y_a$, i.e.,  $L_a(T):= O(\ln T/d(x_a,y_a))$.

For the problem-independent setting, we need to set
$x_a=\mu_a+\Delta_a/3$ and  $y_a=\mu_1-\Delta_a/3$.
This choice implies $\Delta_a^{'2}=(\mu_1-y_a)^{2}=\Delta_a^2/9$. 
Then we can lower bound $d(x_a,\mu_a) \ge 2\Delta_a^2/9$. Thus $L_a(T)= O(\frac{\ln T}{\Delta_a^2}$).
Now by substituting  $\Delta_a$ and $d(x_a,\mu-a)$ in Theorem~\ref{thm:ln} 
for $a \ne 1$, we get 
$\Ex[k_a(T)]\le O(\frac{\ln T}{\Delta_a^2})
$.
Now for arms with $\Delta_a>\sqrt{\frac{N\ln T}{T}}$, we can upper bound the regret by $\Delta_a \Ex[k_a(T)]=O(\sqrt{\frac{T\ln T}{N}})$, and for arms with $\Delta_a\le \sqrt{\frac{N \ln T}{T}}$,  we can upper bound the expected regret by $\sqrt{NT\ln T}$. All in all, it results in the total regret of $O(\sqrt{NT \ln T})$.
\end{proof}

\subsection{Problem-independent Regret Bound with Gaussian Priors}
\label{app:gaussian}

\Gaussian*

The proof is similar to the proof of Theorem~\ref{thm:root} and follows essentially \cite[Theorem 1.3]{agrawal2017near}.
 with Beta priors.
We set $x_a=\mu_a+\Delta_a/3$ and $y_a=\mu_1-\Delta_a/3$.
The lemmas in the previous section for Beta priors hold here with slight modifications. The main lemma that changes for the Gaussian distributions is Lemma~\ref{lem:UPL}.

\begin{lemma} \label{lemma:expdec}
Let $\tau_j$ be the $j$-th time   step in which the optimal arm 1 has been queried. Then 

\[
\Ex\left[\frac{1}{p_{a,\tau_j+1}}-1\right] \le 
\begin{cases}
      e^{11}+5, & \forall j,   \\
      \frac{4}{T\Delta_a^2}, & j>8L_a(t),
\end{cases}
\]
where $L_a(t)=\frac{18\ln(T\Delta_a^2)}{\Delta_a^2}$.
\end{lemma}

\begin{proof}
Note that $p_{a,t}$ is the probability $\Pr(\theta_a(t)>y_a|\mathcal{F}_{B(t)})$. If the prior comes from the Gaussian distribution then $\theta_a(t)$ has distribution $\mathcal{N}(\hat{\mu}_a(t),\frac{1}{k_a(B(t))+1})$.  
Given the definition of $\tau$ and $p_{a,t}$, the proof follows from \citet[Lemma 2.13]{agrawal2017near}. 
\end{proof}

By using Lemma~\ref{lemma:expdec}  and substituting it in eq.~\eqref{eq:pi}, we can   easily obtain the following lemma.
\begin{lemma} \label{lem:e64}
For any arm $a\in [n]$ we have
\[
\sum_{t=1}^{T}\Pr(a(t)=a, E^{\mu}_{a}(t), E_a^{\theta}(t))\le (e^{64}+4)(8L_a(t))+\frac{8}{\Delta_a^2}.
\]
\end{lemma}
\begin{lemma}\label{lem:92delta}
For any arm $a\in [n]$, we have
\[
\sum_{t=1}^{T} \Pr(a(t)=a,\overline{E_a^{\mu}(t)}) \le \frac{1}{d(x_a,y_a)}+1 \le \frac{9}{2\Delta_a^2}+1.
\]
\end{lemma}
Similar to Lemma~\ref{lem:UPL},  we can prove the following lemma.
\begin{lemma}\label{Ldelta}
For any arm $a\in [n]$, we have
\[
\sum_{t=1}^{T} \Pr(a(t)=a,\overline{E_a^{\theta}(t)},E_a^{\mu}(t)) \le L_a(t)+\frac{1}{\Delta_a^2}.
\]
where $L_a(t)=\frac{36\ln(T\Delta_a^2)}{\Delta_a^2}$.

\end{lemma}
\begin{proof}
The proof follows from \cite[Lemma 2.16]{agrawal2017near} and is adapted to the batch setting. 
The decomposition is as in Lemma~\ref{lem:UPL}.
As before, the first term in the decomposition can be upper bounded by $L_a(t)$. 
Instead of bounding the second term with 1, we should bound it with $1/\Delta_a^2$. First, note that
\begin{small}
\[
\sum_{t=1}^{T} \Pr\left(a(t)=a,k_a(B(t))>L_a(t),\overline{E_a^{\theta}(t)},E_a^{\mu}(t)) \right ) \le \Ex\left[\sum_{t=1}^{T} \Pr(\theta_a(t)>y_a|k_a(B(t))>L_a(t),\hat{\mu}_a(B(t))\le x_a ), \mathcal{F}_{B(t)}\right] \ .
\]
\end{small}
We also know that  $\theta_a(t)$ is distributed as $\mathcal{N}(\hat{\mu}_a(t),\frac{1}{k_a(B(t))+1})$. So given $\{\hat{\mu}_a(t)\le x_a\}$, we have that  $\theta_a(t)$ is stochastically dominated by $\mathcal{N}(x_a,\frac{1}{k_a(B(t))+1})$.
Therefore,
\begin{small}
\[
\Pr(\theta_a(B(t))>y_a|k_a(B(t))>L_a(t), \hat{\mu}_a(B(t))\le x_a, \mathcal{F}_{B(t)} ) \le \Pr\left(\mathcal{N}\left(x_a,\frac{1}{k_a(B(t))+1}\right)>y_a | \mathcal{F}_{B(t)}, k_a(B(t))>L_a(t)\right).
\]
\end{small}
By using concentration bounds, we have 
\[
\Pr\left(\mathcal{N}\left(x_a,\frac{1}{k_a(B(t))+1}\right)>y_a\right) \le \frac{1}{2}e^{-\frac{L_a(t)(y_a-x_a)^2}{4}} \le \frac{1}{T\Delta_a^2} \ .
\]
Thus,
\begin{align}
\label{eq:prviol}
\Pr(\theta_a(t)>y_a|k_a(B(t))>L_a(t), \hat{\mu}_a(t)\le x_a, \mathcal{F}_{B(t)} ) \le 1/T\Delta_a^2 \ .
\end{align}

Summing over $t$ will follow the result.
\end{proof}

Using lemmas~\ref{Ldelta},~\ref{lem:e64},~\ref{lem:92delta} we can upperbound
\[
\Ex\left[k_a(t)\right] \le (e^{64}+4) \frac{2 \times 72\ln(T\Delta_a^2)}{\Delta_a^2}+\frac{2\times 4}{\Delta_a^2}+\frac{18\ln(T\Delta_a^2)}{\Delta_a^2}+\frac{1}{\Delta_a^2}+\frac{9}{\Delta_a^2}+1.
\]

Thus, we can upper bound the expected regret due to arm $a$.
Similar to the previous proofs we can upper bound 
\[
\Delta_a \Ex[k_i(T)] \le O\left(\frac{1}{\Delta_a}+\frac{\ln(T\Delta_a^2)}{\Delta_a}\right)+\Delta_a.
\]
Then, if $\Delta_a>e\sqrt{\frac{N\ln N}{T}}$ we can upper bound the regret by $O(\sqrt{\frac{N\ln T}{N}}+1)$.
If $\Delta_a \le e\sqrt{\frac{N\ln N}{T}}$ we can upper bound the regret with $O(\sqrt{NT\ln T})$. Consequently, we can upper bound the total regret by $O(\sqrt{NT \ln T})$ assuming $T\ge N$.

\section{Batch Minimax Optimal Thompson Sampling}\label{sec:B-MOTS-Proof}
In order to increase clarity, we first introduce the main notations used in the proofs. We follow closely the notations used in \cite{MOTS} and adapt them to the batch setting.
\subsection{Notations and Definitions}
Without  loss of generality, we assume the optimal arm is arm $a=1$ with $\mu_1=\max_{a\in\left[N\right]} \mu_a$.
\begin{definition}
Define $\hat{\mu}_{as}$ to be the average reward of arm $a$ when it has been played $s$ times.
\end{definition}

\begin{definition}
We denote by $\mathcal{F}_s$ 
the  history of plays of Algorithm~\ref{alg:MOTS} (B-MOTS) up to the $s$-th pull of arm 1. 
\end{definition}

\begin{definition}
Let $h(j)$ be the largest power of 2 that is less than or equal to $j$.
\end{definition}

\begin{definition}
Define
\[
\mathfrak{B}=\{s=2^i|i=0,\cdots, \log T\} \ .
\]
\end{definition}
We slightly modify ~\citet[eq.(16)]{MOTS} as follows.
\begin{definition}
Define
\begin{equation}
    \Delta=\mu_1 -\min_{s\in \mathfrak{B}} \left\{\hat{\mu}_{1s} + \sqrt{\frac{\alpha}{s} \log^+\left(\frac{T}{sN}\right)} \right\} \ .
\end{equation}
\end{definition}
\begin{definition}\label{def:is}
Similar to the definitions of $D_a(t)$ and $\theta_a(t)$, we define $D_{as}$ as the distribution of arm $a$ when it is played for the $s$-th time.
Also, we define $\theta_{as}$ as a sample from distribution $D_{as}$.
\end{definition}

\begin{lemma} \label{lem:subcon}
Let $X_1, X_2, \cdots$ be independent 1-subgaussian random variables with zero mean. Let's define $\hat{\mu}_t=1/t\sum_{s=1}^t X_s$. Then for $\alpha\ge 4$ and any $\Delta>0$ 
\[
\Pr\left(\exists s\in \mathfrak{B}: \hat{\mu}_s + \sqrt{\frac{\alpha}{s}\log^+(T/sN)} + \Delta \le 0 \right) \le \frac{15N}{T\Delta^2}.
\]
\end{lemma}
The above lemma follows immediately from~\citet[Lemma 9.3]{lattimore2020bandit} as we consider $\mathfrak{B}\subseteq \left[T\right]$. We can strengthen Lemma~\ref{lem:subcon} for Gaussian variables, as described by \citet[Lemma 1]{MOTS} as follows.
\begin{lemma} \label{lem:concJ}
Let $X_a$'s be independent Guassian r.v. with zero mean and variance 1. Denote $\hat{\beta}_t=1/t\sum_{s=1}^t X_s$. Then for $\alpha>2$  and any $\Delta>0$,
\[
\Pr\left(\exists s\in \mathfrak{B}: \hat{\beta}_s + \sqrt{\frac{\alpha}{s} \log^+(T/sN)} + \Delta \le 0 \right) \le \frac{4N}{T\Delta^2} \ . 
\]

\end{lemma}

Now similar to eq.(19) in~\cite{MOTS},  define $\tau_{as}$ as follows. 
\begin{definition}\label{def:tau}
Define
\begin{align}
\tau_{as}=\hat{\mu}_{as}+\sqrt{\frac{\alpha}{s} \log^+(\frac{T}{sN})}   \ .
\end{align}
\end{definition}

\begin{definition}
We define $F_{as}$  as the CDF of distribution for arm $a$ when $k_a(t-1)=s$.
Also $G_{as}(\epsilon)$ is defined as $1-F_{as}(\mu_1-\epsilon)$. \end{definition}
\begin{definition}\label{def:clipped}
Let us define $F'_{as}$ to be the $CDF$ of $\mathcal{N}(\hat{\mu}_{as},1/(\rho s))$.
Moreover, let us define $G'_{as}(\epsilon)= 1-F'_{as}(\mu_1-\epsilon)$. Let $\tilde{\theta}_{as}$ denote a sample from $\mathcal{N}(\hat{\mu}_{as},1/(\rho s))$.
\end{definition}

\begin{definition}\label{def:EB}
Define the event
$E_a(t) = \{ \theta_a(t) \le \mu_1 - \epsilon \}$.
\end{definition}

The following two lemata deal with concentration inequalities that we need for subGaussian random variables.

\begin{lemma}[\citet{MOTS}, Lemma 2]\label{lem:subgaussian}
Let $w>0$ be a constant and $X_1,X_2,\cdots$  be independent and $1$-subGaussian r.v. with zero mean. Denote by $\hat{\mu}_n=\frac{1}{n}\sum_{s=1}^n X_s$. Then for $\alpha>0$ and any $N\le T$,
\[
\sum_{n=1}^T \Pr\left(\hat{\mu}_n+\sqrt{\frac{\alpha}{n}\log^+(N/n)}\ge w\right) \le 1+\frac{\alpha \log^+(Nw^2)}{w^2}+\frac{3}{w^2}+\frac{\sqrt{2\alpha\log^+(Nw^2)}}{w^2}.
\]
\end{lemma}

The following lemma is a variant of \citet[Lemma4]{MOTS}.
\begin{lemma}\label{csaba4}
Let $\rho\in(1/2,1)$ be a constant and $\epsilon>0$. Assuming the reward of each arm is 1-sbuGaussian with mean $\mu_a$. For any fixed $\rho\in(1/2,1)$ and $\alpha>4$,
there exists a constant $c>0$ s.t.
\begin{align}
    \Ex\left[\sum_{s=1}^{T-1}\left(\frac{1}{G'_{1h(s)}(\epsilon)}-1\right)\right] \le \frac{c}{\epsilon^2}.
\end{align}
\end{lemma}
\begin{proof}
The proof closely follows the steps of \citet[Lemma4]{MOTS}. However, for completeness, and for a few differences, we provide the full proof. The main difference is that in Lemma~\ref{csaba4} we have the terms $G'_{1h(s)}$ instead of $G'_{1s}$. We will prove the following two parts:
\begin{itemize}
\item  First, there exists a constant $c'$ such that 
\[
\Ex\left[\frac{1}{G'_{1h(s)}(\epsilon)}-1\right] \le c', \  \forall s,
\]
and 
\item Second, for $L=\left[64/\epsilon^2\right]$, we have
\[
\Ex\left[\sum_{s=L}^{T}(\frac{1}{G'_{1h(s)}(\epsilon)}-1)\right] \le \frac{4}{e^2}(1+\frac{16}{\epsilon^2}) \ .
\]
\end{itemize}
Denote by  $\Theta_s=\mathcal{N}(\hat{\mu}_{1h(s)},1/(\rho h(s)))$.
Also, let $Y_s$ be the number of trials until a sample from $\Theta_s$ becomes greater than $\mu_1-\epsilon$. By the definition of $G'_{ah(s)}$ we have
\[
\Ex\left[\frac{1}{G'_{1h(s)}(\epsilon)}-1\right]=\Ex\left[Y_s\right].
\]
Similar to \cite[Eq. (59)]{MOTS} one can show that
\[
\Pr(Y_s<r) \ge 1-r^{-2}-r^{-\frac{\rho'}{\rho}} \ .
\]
Define $z=\sqrt{2\rho' \log r}$, for $r\ge 1$, where $\rho'\in (\rho,1)$. 
Also let $M_r$ be the maximum of $r$ independent samples from $\Theta_s$. Thus
\begin{align*}
\Pr(Y_s < r) \ge& \Pr(M_r>\mu_1-\epsilon)\\
\ge& \Ex\left[\Ex\left[\mathbb{I}(M_r>\hat{\mu}_{1h(s)}+\frac{z}{\sqrt{\rho h(s)}}, \hat{\mu}_{1h(s)}+\frac{z}{\sqrt{\rho h(s)}}
\ge \mu_1-\epsilon)\right]|\mathcal{F}_{h(s)}\right]\\
=& \Ex\left[\mathbb{I}(\hat{\mu}_{1h(s)}+\frac{z}{\sqrt{\rho h(s)}} \ge \mu_1-\epsilon)\times\Pr\left(M_r > \hat{\mu}_{1h(s)}+\frac{z}{\sqrt{\rho h(s)}}| \mathcal{F}_{h(s)}\right)\right] \ .
\end{align*}
For a random variable $Z\sim \mathcal{N}(\mu,\sigma^2)$  we have the following tail bound
\[
\Pr(Z>\mu+x\sigma) \ge \frac{1}{\sqrt{2\pi}}\frac{x}{x^2+1} e^{-\frac{x^2}{2}} \ .
\]
Thus, for $r>e^2$, 
\[
\Pr\left(M_r>\hat{\mu}_{1h(s)}+\frac{z}{\sqrt{\rho h(s)}}|\mathcal{F}_{h(s)}\right) \ge 1-\exp\left(-\frac{r^{1-\rho'}}{\sqrt{8\pi \log r}}\right) \ .
\]
Similar to \citet{MOTS}, we can show that if $r\ge \exp(10/(1-\rho')^2)$ we have 
\[
\Pr\left(M_r> \hat{\mu}_{1h(s)}+\frac{z}{\sqrt{\rho h(s)}} | \mathcal{F}_{h(s)}\right) \ge 1-\frac{1}{r^2} \ . 
\]
Also, for $\epsilon>0$, we have
\[
\Pr\left(\hat{\mu}_{1h(s)}+\frac{z}{\sqrt{\rho h(s)}} \ge \mu_1-\epsilon\right) \ge 1-r^{-\rho'/\rho} \ .
\]
Therefore, for $r\ge \exp(10/(1-\rho')^2)$, we obtain
\[
\Pr(Y_s<r) \ge 1-r^{-2}-r^{-\rho'/\rho} \ .
\]
For any $\rho' >\rho$ we get
\[
\Ex\left[Y_s\right] =\sum_{r=0}^{\infty} \Pr(Y_s\ge r) \le 2\exp\left(\frac{10}{(1-\rho')^2}\right) + \frac{1}{(1-\rho)-(1-\rho')} \ .
\]
By setting $1-\rho'=(10\rho)/2$,
\[
\Ex\left[\frac{1}{G'_{1h(s)}(\epsilon)}-1\right] \le 2\left(\frac{40}{(1-\rho)^2}\right)+\frac{2}{1-\rho} \ .
\]
Now because $\rho$ is fixed, there exists a universal constant $c'>0$  s.t.
\[
\Ex\left[\frac{1}{G'_{1h(s)}(\epsilon)}-1\right] \le c' \ .
\]
Proof of the second part is similar.
\end{proof}
In the above proof, we had to be careful about the conditional expectations as the history in the batch mode, namely, $\mathcal{F}_{h(s)}$, is different from the sequential setting $\mathcal{F}_{s}$. Apart from that, as we stated, the proof is identical to \citet[Lemma4]{MOTS}.

\subsection{Clipped Gaussian Distribution}
\label{app:BMOTS}

\mots*

\begin{proof}
We closely follow the proof of of the fully sequential algorithm, provided in \citet[Theorem 1]{MOTS}, and adapt it to the batch setting. Let us define
\[
S:=\{a:\Delta_a > \max\{2\Delta,8\sqrt{N/T}\}\} \ .
\]
Then, as \citet[eq. (17)]{MOTS} argued, we have
\begin{align}\label{eq:de}
    \mathcal{R}(T) &\le  \sum_{a:\Delta_a>0} \Delta_a \Ex\left[k_a(t)\right] \nonumber \\
    &\le \Ex\left[2T\Delta\right] + 8\sqrt{NT} + \Ex\left[\sum_{a\in S} 
    \Delta_a k_a(t)\right] \ .
\end{align}
where as in  \citet[eq. (18)]{MOTS} (which immediately follows from Lemma~\ref{lem:concJ}) we have $\Ex\left[2T\Delta\right] \le4/\sqrt{15NT}$.
By Definition~\ref{def:tau}, we have $\tau_{as}=\tau_a(t)$ when $k_a(t)=s$. Thus, for $a\in S$, we get
  \[
  \tau_{1s}\ge \mu_1- \Delta \ge \mu_1-\frac{\Delta_a}{2}.
  \]
Therefore, for $\tilde{\theta}_{is}$ as defined in the definition~\ref{def:clipped}, we have
\[
\Pr(\tilde{\theta}_{1s}\ge \mu_1-\Delta_a/2)=\Pr(\theta_{1s}\ge \mu_1-\Delta_a/2).
\]
Hence for $a\in S$, we have 
$$G_{1s}(\Delta_a/2)=G'_{1s}(\Delta_a/2).   $$

 For Algorithm~\ref{alg:MOTS}, we need to revise Theorem 36.2 in~\cite{lattimore2020bandit} as follows.
Note that we start from $t=N+1$ and $s=1$ since the algorithm plays each arm once in the beginning.
\begin{lemma}\label{lemma:book}
For $\epsilon >0$, the expected number of times Algorithm~\ref{alg:MOTS} plays arm $a$ is bounded by

\begin{align}
\Ex\left[k_a(t)\right] \le& \Ex\left[\sum_{t=1}^T \mathbb{I}\{a(t)=a,E_a(t)\}\right] + \Ex\left[\sum_{t=1}^T \mathbb{I}\{a(t)=a,\overline{E_a}(t)\}\right]  \nonumber \\
\le& 1+ \Ex\left[\sum_{t=0}^{T-1}  \left(\frac{1}{G_{1k_1(p_1(t))}}-1\right)\mathbb{I}\{a(t)=1\}\right] + \Ex\left[\sum_{t=N+1}^{T-1} \mathbb{I} \{a(t)=a, \overline{E_a(t)} \}\right] \\
\le& 2+ 
\Ex\left[\sum_{s=0}^{T-1} \left(\frac{1}{G_{1h(s)}(\epsilon)}-1\right)\right] 
+ \Ex\left[\sum_{s=0}^{T-1} \mathbb{I} \{G_{ah(s)}(\epsilon)>1/T\}  \right] \ .
\end{align}

\end{lemma}
\begin{proof}
We follow the steps in~\cite{lattimore2020bandit} and make appropriate  modifications for  our batch mode algorithm. As defined in Definition~\ref{def:EB}, $E_a(t) = \{ \theta_a(t) \le \mu_1 - \epsilon \}.$
Thus,
\[
\Pr(\theta_1(t) \ge \mu_1 -\epsilon | \mathcal{F}_{B(t)}) = G_{1k_1(B(t))} \ .
\]
Now we consider the following decomposition based on $E_a(t)$ as follows,
\begin{align}\label{eq:dec}
\Ex\left[k_a(t)\right] = \Ex\left[\sum_{t=1}^{T} \mathbb{I} \{a(t)=a, E_a(t)\}\right] +  \Ex\left[\sum_{t=1}^{T} \mathbb{I} \{a(t)=a, \overline{E_a(t)}\}\right] \ .
\end{align}
An upper bound for the first terms is as follows. Let $a'(t)=\argmax_{a\ne 1} \theta_a(t)$. Then,
\begin{align*}
\Pr(a(t)=1, E_a(t) | \mathcal{F}_{B(t)}) &\ge \Pr(a'(t)=a,E_a(t),\theta_1(t)\ge \mu_1-\epsilon| \mathcal{F}_{B(t)}) \\
&= \Pr(\theta_1(t)\ge \mu_1-\epsilon | \mathcal{F}_{B(t)}) \Pr(a'(t)=a,E_a(t)|\mathcal{F}_{B(t)})\\
&\ge \frac{G_{1k_1(B(t))}}{1-G_{1k_1(B(t))}} \Pr(a(t)=a,E_a(t) | \mathcal{F}_{B(t)}) \ .
\end{align*}
In the first equality, we use the fact that  $\theta_1(t)$ is conditionally independent of $a'(t)$ and $E_a(t)$, given $\mathcal{F}_{B(t)}$.
For the second inequality we use
\[
\Pr(a(t)=a,E_a(t)|\mathcal{F}_{B(t)}) \le (1-\Pr(\theta_1(t)>\mu_1-\epsilon|\mathcal{F}_{B(t)}))\Pr(a'(t)=a,E_a(t)|\mathcal{F}_{B(t)}) \ .
\]
Therefore,
\[
\Pr(a(t)=a,E_a(t)|\mathcal{F}_{B(t)}) \le \left(\frac{1}{G_{1k_1(B(t))}}-1\right) \Pr(a(t)=1|\mathcal{F}_{B(t)}) \ .
\]
By substituting this into \eqref{eq:dec}, we obtain
\begin{align*}
\Ex\left[\sum_{t=1}^{T} \mathbb{I} \{a(t)=a, E_a(t) \}\right] \le& \Ex\left[\sum_{t=1}^{T} (\frac{1}{G_{1k_1(B(t))}}-1) \mathbb{I}\{a(t)=1\}|\mathcal{F}_{B(t)}\right]\\
=&
\Ex\left[\sum_{t=1}^{T}(\frac{1}{G_{1k_1(B(t))}}-1) \mathbb{I}(a(t)=1)\right]\\
\le& \Ex\left[\sum_{s=0}^{T-1} (\frac{1}{G_{1h(s)}}-1)\right]  \ .
\end{align*}

Now define
\[
\tau=\{t\in \left[T\right]:1-F_{a k_a(B(t))} (\mu_1-\epsilon)>1/T\} \ .
\]

For the second expression in \eqref{eq:dec},
 we get 
\begin{align*}
\Ex\left[\sum_{t=1}^T \mathbb{I}(a(t)=a, \overline{E_a(t)})\right] \le & \Ex\left[\sum_{t\in \tau} \mathbb{I}(a(t)=a)\right]+\Ex\left[\sum_{t\notin \tau}\mathbb{I}(\overline{E_a(t)})\right] \\
&\le  \Ex\left[\sum_{s=0}^{T-1}\mathbb{I}\{1-\mathcal{F}_{ah(s)}(\mu_1-\epsilon)\}>1/T\right]+\Ex\left[\sum_{t\notin \tau} \frac{1}{T}\right] \\
&\le \Ex\left[\sum_{s=0}^{T-1}\mathbb{I}(G_{ah(s)}>1/T)\right]+1.
\end{align*}
\end{proof}
Now by setting $\epsilon=\Delta_a/2$ we can show that 
\begin{align}\label{eq:decomposition}
\Delta_a \Ex\left[k_a(t)\right]  \le  \Delta_a + \Delta_a  \Ex\left[\sum_{N+1}^{T-1} \mathbb{I}\{ a(t)=a, \overline{E_a(t)}\}\right] + \Delta_a \Ex\left[\sum_{t=1}^{T-1} (\frac{1}{G'_{1k_1(B(t))}(\Delta_a/2)}-1) \mathbb{I}(a(t)=1)\right].
\end{align}
To bound  the first term
we note that

\[
\overline{E_a(t)} \subseteq \left\{ \hat{\mu}_a(B(t))+\sqrt{\frac{\alpha}{k_a(B(t))} \log^+\left(\frac{T}{Nk_a(B(t))}\right) } > \mu_1- \Delta_a/2 \right\} \ .
\]
Define $\kappa_a$  as the sum of the event in the right hand side of the above equation, namely,
\begin{align}
\kappa_a=\sum_{s=1}^{T} \mathbb{I}\left\{\hat{\mu}_{ah(s)}+\sqrt{\alpha/h(s) \log^+(T/h(s)N)}>\mu_1-\frac{\Delta_a}{2}\right\} \ .
\end{align}
Hence,
\[
\Delta_a  \Ex\left[\sum_{N+1}^{T-1} \mathbb{I}\{ a(t)=a, \overline{E_a(t)}\}\right]  \le \Delta_a \Ex\left[\kappa_a\right] =\Delta_a \Ex\left[\sum_{s=1}^{T} \mathbb{I} \left\{\hat{\mu}_{ah(s)}+\sqrt{\frac{\alpha}{h(s)}\log^+\left(\frac{T}{h(s)N}\right)  } > \mu_1 - \Delta_a/2\right\}\right].
\]

Using

Lemma~\ref{lem:subgaussian} 
and the fact that $\Delta_a=\mu_1-\mu_a$ we have
\begin{align}
\Delta_a\Ex\left[\kappa_a\right] &\le \Delta_a \sum_{s=1}^{T} \Pr\left\{\hat{\mu}_{ah(s)}-\mu_a+\sqrt{\frac{\alpha}{h(s)}} \log^+(T/h(s)N)>\frac{\Delta_a}{2}\right\} \\
&\le \Delta_a+\frac{12}{\Delta_a}+\frac{4\alpha}{\Delta_a} \left( \log^+(\frac{T\Delta_a^2}{4N}) + \sqrt{2\alpha \pi \log^+(\frac{T\Delta_a^2}{4N})} \right) \ .
\end{align}
Now it implies that $\Ex\left[\Delta_a \kappa_a\right] = O(\sqrt{T/k}+\Delta_a)$. 
For bounding the second term of \eqref{eq:decomposition}, a slight modification of 
Lemma~\ref{csaba4}, provides 

\[
\Delta_a \Ex\left[\sum_{t=1}^{T-1} \left(\frac{1}{G'_{1k_1(B(t))}(\Delta_a/2)}-1\right) \mathbb{I}(a(t)=1)\right] = \Delta_a \Ex\left[\sum_{s=1}^{T-1}\left(\frac{1}{G'_{1h(s)}(\Delta_a/2)}-1\right)\right] = O(\sqrt{T/N}) \ .
\]

\end{proof}

\subsection{MOTS 1-subgaussian asymptotic regret bound}
\label{app:assBMOTS}

\assymp*

First we should prove the following lemma, which a simple variant of \citet[Lemma 6]{MOTS} for the batch setting.  
\begin{lemma} \label{lem:rhoexp}
For any $\epsilon_T>0$, and $\epsilon>0$ that satisfies $\epsilon+\epsilon_T < \Delta_a$, it holds that
\[
\Ex\left[\sum_{s=1}^{T-1} \mathbb{I}\{G'_{ah(s)}>1/T \}  \right] \le 1+\frac{4}{\epsilon_T^2} + \frac{4\log T}{\rho(\Delta_a-\epsilon-\epsilon_T)^2} \ .
\]
\end{lemma}
\begin{proof}
The proof closely follows \citet[Lemma 6]{MOTS} and adapted to the batch setting. As before $\mu_a+\epsilon_T \le \mu_1 -\epsilon$, and by using the tail-bound for $\sigma$-subGaussian random variables  we have
\[
\Pr(\hat{\mu}_{ah(s)} > \mu_a+\epsilon_T) \le \exp(-h(s)\epsilon_T^2/2) \le \exp(-s\epsilon_T^2/4).
\]
Furthermore 
\[
\sum_{s=1}^{\infty} \exp\left(-\frac{s\epsilon_T^2}{4}\right) \le 4/\epsilon_T^2.
\]
Define 
\[
L_a=4\log T/(\rho(\Delta_a-\epsilon-\epsilon_T)^2). 
\]

For $s\ge L_a$, let $X_{as}$ be sampled from $\mathcal{N}(\hat{\mu}_{ah(s)},1/(\rho h(s)))$.
Then if we have $\hat{\mu}_{ah(s)} \le \mu_a + \epsilon_T$, the Guassian tail bound  implies 
\[
\Pr(X_{as} \ge \mu_1-\epsilon ) \le \frac{1}{2} \exp\left(-\frac{\rho h(s)(\Delta_a-\epsilon-\epsilon_T)^2}{2}\right) \le 1/T \ .
\]
Now, denote the event $\{\hat{\mu}_{ah(s)} \le \mu_a+\epsilon_T\}$  by $Y_{as}$. By using the fact that $\Pr(A)\leq \Pr(A|B)+1-\Pr(B)$, we have

\begin{align*}
\Ex\left[\sum_{s=1}^{T-1} \mathbb{I}\{G'_{ah(s)}(\epsilon)>1/T\}\right] &=\sum_{s=1}^{T-1} \Pr(\{G'_{ah(s)}(\epsilon)>1/T\})\\
&\leq \sum_{s=1}^{T-1} \Pr(\{G'_{ah(s)}(\epsilon)>1/T\}|Y_{as}) + \sum_{s=1}^{T-1}(1-\Pr(Y_{as}))\\
&\le \lceil L_a\rceil+\sum_{s=1}^{T-1}(1-\Pr(Y_{as}))\\ 
&\le 1+\frac{4}{\epsilon_T^2} + \frac{4\log T}{\rho(\Delta_a-\epsilon-\epsilon_T)^2} \ .
\end{align*}

\end{proof}
Now, closely following the proof of ~\citet[Theorem 2]{MOTS}, we define
\begin{align}\label{eq:zeps}
Z(\epsilon)= \left\{ \forall s\in \mathfrak{B} : \hat{\mu}_{1s}+\sqrt{\frac{\alpha}{s}\log^+(\frac{T}{sN})} \ge \mu_1-\epsilon \right\} \ .
\end{align}
For an arm $a\in\left[N\right]$, we have
\begin{align*}
    \Ex\left[k_a(t)\right]\le& \Ex\left[k_a(t)|Z(\epsilon)\right]\Pr(Z(\epsilon))+T(1-\Pr(Z(\epsilon)))\\
    \le& 2+\Ex\left[\sum_{s=1}^{T-1}(\frac{1}{G_{1h(s)}(\epsilon)}-1)|Z(\epsilon)\right] + T(1-\Pr(Z(\epsilon))) + \Ex\left[\sum_{s=1}^{T-1} \mathbb{I}(G_{ah(s)}(\epsilon)>1/T)\right]\\
    \le& 2+\Ex\left[\sum_{s=1}^{T-1}(\frac{1}{G'_{1h(s)}(\epsilon)})\right] + T(1-\Pr(Z(\epsilon))) + \Ex\left[\sum_{s=1}^{T-1}\mathbb{I}(G'_{ah(s)}(\epsilon)>1/T)\right] \ .
\end{align*}
The second inequality is due to  Lemma~\ref{lemma:book} and the last inequality is due to the  fact that given $Z(\epsilon)$, we have $G_{1h(s)}(\epsilon)=G'_{1h(s)}(\epsilon)$. Also, note that if 
\[
\hat{\mu}_{ah(s)}+ \sqrt{\frac{\alpha}{h(s)}\log^+(T/h(s)N)} \ge \mu_1 -\epsilon,
\]
then we have $G_{ah(s)}(\epsilon) = G'_{ah(s)}(\epsilon)$, or otherwise we have $G_{ah(s)}(\epsilon) =0 \le G'_{as}(\epsilon)$. 

Now from Lemma~\ref{lem:subcon}
and by setting $\epsilon=\epsilon_T=\frac{1}{\log\log T}$, we have
\[
T(1-\Pr(Z(\epsilon))) \le 15N(\log \log T)^2.
\]
By using Lemma~\ref{csaba4}
\[
\Ex\left[\sum_{s=1}^{T-1}(\frac{1}{G'_{1h(s)}(\epsilon)}-1)\right] \le O( (\log \log T)^2) \ .
\]
Then, by Lemma~\ref{lem:rhoexp}
\[
\Ex\left[\sum_{s=1}^{T-1} \mathbb{I}(G'_{ah(s)}(\epsilon)>1/T)\right] \le 1 + 4(\log \log T)^2 + \frac{4\log T}{\rho(\Delta_a-2/\log\log T)^2} \ .
\]
The theorem will follow easily by combining the above equations, namely,
\[
\lim_{T\rightarrow \infty} \frac{\Ex\left[\Delta_a k_a(t)\right]}{\log T} = \frac{2}{\rho \Delta_a} \ .
\]
\subsection{MOTS for Gaussian Rewards}
\label{app:BMOTSJ}

\motsj*

Recall that
$F'_{as}$ denotes the CDF of 
$\mathcal{J}(\hat{\mu}_{as},1/s)$ for any $s\ge 1$
and $G'_{as}=1-F'_{as}(\mu_1-\epsilon)$. 
We closely follow the recipe of \cite[Theorem 4]{MOTS}.  
The proof of the minimax   and asymptotic-optimal bounds are similar to the proof of Theorem~\ref{thm:minimax} and ~\ref{thm:asymptotic} with a few differences. Note that in the proof of Theorem~\ref{thm:minimax},
we used the fact that $\rho<1$ (used in the definition of the Gaussian distribution $\tilde{\theta}_a$). In  Theorem~\ref{thm:MOTS-J}, we do not have the parameter $\rho$.
Therefore instead of 
Lemma~\ref{csaba4}
we prove the following, which is a batch variant of \citet[Lemma 9]{MOTS}.

\begin{lemma} \label{lemma:norho}
There exists a universal constant $c$, s.t.,
\[
\Ex\left[\sum_{s=1}^{T-1} (\frac{1}{G'_{1h(s)}(\epsilon)}-1)\right] \le c/\epsilon^2 \ .
\]

\end{lemma}
\begin{proof}
Similar to 
(Lemma~\ref{csaba4}), the following two statements need to be proven:
\\
(i) there exists a universal constant $c'$ s.t. 
\[
\sum_{s=1}^{L} \Ex\left[\frac{1}{G'_{1h(s)}(\epsilon)}-1\right] \le \frac{c'}{\epsilon^2}, \forall s \ .
\]
(ii) for $L=\left[64/\epsilon^2\right]$ 
\[
\Ex\left[\sum_{s=L}^T (\frac{1}{G'_{1h(s)}(\epsilon)}-1)\right] \le \frac{4}{e^2} (1+16/\epsilon^2) \ .
\]
The proof of statement (ii) is similar to the one in Lemma~\ref{lem:concJ}. Therefore, 
We  focus on the first statement here, which closely follows the proof of \citet[Lemma 9]{MOTS}.

Let $\hat{\mu}_{1h(s)}=\mu_1+x$. Let $Z$ be a sample from $\mathcal{J}(\hat{\mu}_{1h(s)},1/h(s))$. For $x<-\epsilon$, applying Lemma~\ref{lem:concJ} with $z=-\sqrt{h(s)}(\epsilon+x) > 0$ we have

\begin{equation} \label{eq:conc}
G'_{1h(s)} (\epsilon) = \Pr(Z>\mu_1-\epsilon) = \frac{1}{2} \exp\left(-\frac{h(s)(\epsilon+x)^2}{2} \right) \ .
\end{equation}
Note that $x \sim \mathcal{N}(0,1/h(s))$. Let $f(x)$ be the PDF of $\mathcal{N}(0,1/h(s))$.

\begin{align*}
    \Ex_{x\sim \mathcal{N}(0,1/h(s))} \left[\left(\frac{1}{G'_{1h(s)}(\epsilon)}-1 \right)\right]=&\int_{\infty}^{-\epsilon} f(x) \left(\frac{1}{G'_{1h(s)}(\epsilon)}-1\right) dx+\int_{-\epsilon}^{-\infty} f(x)\left(\frac{1}{G'_{1h(s)}(\epsilon)}-1\right) dx\\
    \le& \int_{-\infty}^{-\epsilon} f(x)\left(2\exp\left(\frac{h(s)(\epsilon+x)^2}{2}\right)-1\right) dx+ \int_{-\epsilon}^{\infty} f(x) \left(\frac{1}{G'_{1h(s)}}(\epsilon)-1\right) dx\\
    \le& \int_{-\infty}^{-\epsilon} f(x)\left(2\exp\left(\frac{h(s)(\epsilon+x)^2}{2}\right)-1\right) dx+ 
    \int_{-\epsilon}^{\infty} f(x)  dx\\
    \le& \sqrt{2}\frac{e^{-s\epsilon^2/4}}{\sqrt{s}\epsilon}+1
\end{align*}
The first inequality is because of eq.~\eqref{eq:conc}. The second inequality is because $G'_{1h(s)}(\epsilon)=\Pr(Z>\mu_1-\epsilon)\ge 1/2$, since $\hat{\mu}_{1h(s)}=\mu_1+x\ge \mu_1-\epsilon$. And the last inequality is due to the definition of $h(s)$.

Also for $s\le L$, we have $e^{-s\epsilon^2/4}=O(1)$, thus for $L=\left[\frac{64}{\epsilon^2}\right]$,
\[
\sum_{s=1}^{L}\Ex\left[\left(\frac{1}{G'_{1h(s)}(\epsilon)} -1\right)\right] = O\left(\sum_{s=1}^{L}\frac{1}{\sqrt{s}\epsilon}\right)=O(1/\epsilon^2) \ .
\]

\end{proof}
From the above lemma we have
\[
\Delta_a \Ex\left[\sum_{s=1}^{T-1}\left(\frac{1}{G'_{1h(s)}}(\Delta_a/2)-1\right)\right] \le O(\sqrt{T/K}+\Delta_a).
\]
The rest of the proof for minimax optimality is similar to the proof of Theorem~\ref{thm:minimax}.

For the asymptotic regret bound, we first state the following lemma, which the batch mode version of 
\begin{lemma}\label{lem:B18}
 for any $\epsilon_T>0, \epsilon>0$ that satisfies $\epsilon+\epsilon_T<\Delta_a$, we have
 \[
 \Ex\left[\sum_{s=1}^{T-1} \mathbb{I}\{G'_{ih(s)}>1/T\}\right] \le 1+\frac{4}{\epsilon_T^2} + \frac{4\log T}{(\Delta_a -\epsilon-\epsilon_T)^2}.
 \]
 \end{lemma}
 \begin{proof}
 The proof is similar to the proof of Lemma~\ref{lem:rhoexp}.
 \end{proof}

The  proof asymptotic regret bound  is similar to the proof of Theorem~\ref{thm:asymptotic} where we use Lemmas~\ref{lemma:norho},~\ref{lem:concJ}, and \ref{lem:B18}. 

\section{Batch Thompson Sampling for Contextual Bandits}
First, we reintroduce a number of  notations from \cite{contextual} and adapt them to the batch setting. 
\subsection{Notations and Definitions}
In time step $t$ of the B-TS-C algorithm, we generate a sample $\tilde{\mu}(t)$ from
$\mathcal{N}(\hat{\mu}(B(t)),v^2{\cal B}(B(t))^{-1})$ 
and play the arm $a$ with maximum $\theta_a(t)=b_a(t)^T\tilde{\mu}(t)$.

\begin{definition}\label{def:stand}
Let us define the standard deviation of empirical mean in the batch setting as 
\[
s_{a}(B(t)):=\sqrt{b_a(t)^T {\cal B}(B(t))^{-1}b_a(t)}.
\]
\end{definition}
\begin{definition}
Let us define the history of the process up to time $t$ by $$H_{t} =\{a(\tau),r_{a(\tau)}(\tau),b_a(\tau)| a\in[N],\tau\in[t]\},$$
where $a(\tau)$ indicates the arm played at time $\tau$, $b_a(\tau)$ indicates the context vector associated with arm $a$ at time $\tau$, and $r_{a(\tau)}$ indicates the reward at time $\tau$.
\end{definition}

\begin{definition}
Define the filtration $\mathcal{F}_{B(t)}$ as the union of history until time $B(t)$, and the context vectors up to time $t$, i.e., 

\[
\mathcal{F}_{B(t)} = \{H_{B(t)},b_a(t') | a\in[N], t'\in(B(t),t] \}.
\]

\end{definition}

\begin{definition}\label{def:noise}
We assume that $\eta_{a,t}  = r_{a}(t) - \langle b_a(t), \mu \rangle$, conditioned on $\mathcal{F}_{B(t)}$,  is $\sigma$-subGaussian for some $\sigma\geq 0$.
\end{definition}

\begin{definition}
Define 
\begin{align*}
l(t)&=\sigma\sqrt{d\ln\frac{t^3}{\delta}}+1,\\
v(t) & = \sigma\sqrt{9d\ln\frac{t}{\delta}},\\
p &= \frac{1}{4e\sqrt{\pi}},\\ 
g(t)&=\min\{\sqrt{4d\ln(t)}, \sqrt{4\log(tN)}\}v(t)+l(t).
\end{align*}
\end{definition}
\begin{definition}
Define $E^{\mu}(t)$ as the event that for any arm $a$
\[
\left \{ |\langle b_a(t),\hat{\mu}(B(t))-b_a(t)^{\top} \mu\rangle|\le l(t) s_{a}(B(t)) \right \}.
\]
\end{definition}

\begin{definition}
Define $E^{\theta}(t)$ as the event
\[
\left \{ \forall a:|\theta_a(t)-\langle b_a(t),\hat{\mu}(B(t))\rangle| \le (g(t)-l(t))s_{a}(B(t)) \right \}.
\]
\end{definition}
\begin{definition}
Define the difference between the mean reward of the optimal arm at time $t$, denoted by $a^*(t)$, and arm $a$ as follows
\[
\Delta_a(t) = \langle b_{a^*(t)}(t), \mu \rangle - \langle b_{a}(t), \mu \rangle \ .
\]
\end{definition}

\begin{definition}\label{def:saturated}
We say that an arm is saturated at time $t$ if $\Delta_a(t)> g(t) s_{a}(B(t))$. We also denote by $C(t)$ the set of saturated arms at time $t$. An arm $a$ is unsaturated at time $t$ of $a\notin C(t)$.

\end{definition}

\begin{lemma}[\cite{abbasi2011improved}]\label{lem:abbasi}
Let $\mathcal{F}'_t$ be a filteration. Consider two random processes $m_t\in\mathbb{R}^d$ and $\mu_t\in\mathbb{R}$ where $m_t$ is $\mathcal{F}'_{t-1}$-measurebale and $\mu_t$ is a martingale difference process and  $\mathcal{F}'_t$-measurebale. Define, $\xi_t= \sum_{\tau=1}^t m_{\tau}\mu_t$ and $M_t = I_d + \sum_{\tau=1}^{t} m_{\tau}m_{\tau}^{\top}$. Assume that given $\mathcal{F}_t'$, $\mu_t$ is $\sigma$-subGaussian. Then, with probability $1-\delta$, 
$$\|\xi_t\|_{M_t^{-1}}\leq \sigma \sqrt{d\ln\frac{t+1}{\delta}}.$$
\end{lemma}

\subsection{Analysis}
\label{app:CB}

\contextual*

The proof closely follows ~\cite[Theorem 1]{contextual}.
We first start with the  following lemma, that is a batch version of ~\cite[Lemma 1]{contextual}.

\begin{lemma} \label{lemma:mutheta}
For all $t$, and $0<\delta<1$, we have $\Pr(E^{\mu}(B(t)))\ge 1-\delta/t^2 $. Moreover,
For all filtration $\mathcal{F}_{B(t)}$, we have $\Pr(E^{\theta}(t)|\mathcal{F}_{B(t)})\ge 1-1/t^2$.
\end{lemma}
\begin{proof} The proof closely follows \citet[Lemma 1]{contextual} where we adapt it to the batch setting. We only prove the first part as the second part very similar. We first invoke Lemma~\ref{lem:abbasi} as follows.  Set $m_t=b_{a(t)}(t)$, $\eta_t=r_{a(t)}(t)-b_{a(t)}(t)^T \mu,$
and
\[
F'_{t}=\{a(\tau+1), m_{\tau+1}:\tau\le t\} \cup \{\eta_{\tau}:\tau \le B(t) \} .
\]

Note that  $\eta_{t}$ is conditionally $\sigma$-subgaussian, and is a martingale difference process. Therefore,
\[
\Ex\left[\eta_t|F'_{B(t)}\right]=\Ex\left[r_{a(t)}|b_{a(t)}(t),a(t)\right]- \langle b_{a(t)}(t), \mu \rangle=0 \ .
\]
Thus, we have
\[
M_t=I_d+\sum_{\tau=1}^{t} m_{\tau} m_{\tau}^{\top} \
\]
and
\[
\xi_t=\sum_{\tau=1}^{t} m_{\tau} \eta_{\tau} \ .
\]
Similar to  \citet[Lemma 1]{contextual}, we have ${\cal{B}}(t)=M_{t-1}$, but we need to change  $\hat{\mu}(t)-\mu=M_{B(t)}^{-1}(\xi_{B(t)}-\mu)$. 
For any vector $y\in \mathbb{R}$ and matrix $A\in \mathbb{R}^{d\times d}$, let us define the norm $\|y\|_{A}:=\sqrt{y^T A y}$. Hence, for all $a$,
\[
|\langle b_a(t),\hat{\mu}(t)\rangle - \langle b_a(t), \mu \rangle|= \|b_a(t)\|_{{\cal{B}}(t)^{-1}} \times \|\xi_{B(t)}-\mu\|_{M_{B(t)}^{-1}}.
\]
Since $B(t)\leq t-1$, Lemma~\ref{lem:abbasi} implies that with probability at least $1-\delta'$,
\[
\|\xi_{B(t)}\|_{M_{B(t)}^{-1}} \le \sigma \sqrt{d \ln(t/\delta')}.
\]
Thus,
\[
\|\xi_{B(t)}-\mu\|_{M_{B(t)}^{-1}} \le \sigma\sqrt{d\ln(t/\delta')} + \|\mu\|_{M_{B(t)}^{-1}} \le \sigma \sqrt{d \ln(t/\delta')}+1.
\]
Now by setting $\delta'=\frac{\delta}{t^2}$ we have with probability $1-\delta/t^2$, and for all arms $a$,
\[
|\langle b_a(t), \hat{\mu}(B(t)) \rangle -\langle b_a(t), \mu \rangle| \le l(t) s_{a}(B(t)) \ .
\]
\end{proof}
Now, we lower bound the probability that $\theta_{a^*(t)}(t)$ becomes larger than $\langle b_{a^*(t)}(t), \mu \rangle$.
\begin{lemma}\label{lem:prteta}
For any filtration $\mathcal{F}_{B(t)}$, if $E^{\mu}(t)$ holds true, we have
\[
\Pr\left(\theta_{a^{*}(t)}(t) > \langle b_{a^*(t)}(t),  \mu \rangle|\mathcal{F}_{B(t)}\right) \ge p.
\]

\end{lemma}
\begin{proof}
The proof easily follows from \citet[Lemma 2]{contextual}.
Suppose $E^{\mu}(t)$ holds true, then 
\[
|\langle b_{a^*(t)}(t), \hat{\mu}(t) \rangle- \langle b_{a^*(t)}(t), \mu \rangle| \le \ell(t) s_{a^*(t)}(B(t)) \ .
\]
The Gaussian random variable $\theta_{a^*(t)}(t)$ has mean $\langle b_{a^*(t)}(t), \hat{\mu}(t) \rangle$ and standard deviation $v_ts_{a^*(t)}(B(t))$. Therefore, we have
\[
\Pr(\theta_{a^*(t)}(t) \ge \langle b_{a^*(t)}(t), \mu \rangle|\mathcal{F}_{B(t)}) \ge \frac{1}{4\sqrt{\pi}}e^{-Z_t^2} \ .
\]
where $|Z_t| = \left|\frac{\langle b_{a^*(t)}(t), \hat{\mu}(t) \rangle- \langle b_{a^*(t)}(t), \mu \rangle}{v(t) s_{a^*(t)}(B(t))}\right| \le 1$.
\end{proof}
The following lemma bounds the probability that an arm played at time $t$ is not saturated. 
\begin{lemma}\label{lemma:pminus}
Given $\mathcal{F}_{B(t)}$,  if $E^{\mu}(t)$ is true,
\[
\Pr(a(t)\notin C(t) |\mathcal{F}_{B(t)})
\ge p-\frac{1}{t^2} \ .
\]
\end{lemma}
\begin{proof}
The proof is a slight modification of \citet[Lemma 3]{contextual} for the batch setting.
If $\forall j\in C(t)$ we have $\theta_{a^*(t)}(t) > \theta_j(t)$, then one of the unsaturated actions much be played which leads us to
\[
\Pr(a(t)\notin C(t)|\mathcal{F}_{B(t)}) \ge \Pr(\theta_{a^*(t)}(t)>\theta_j(t), \forall j\in C(t)|\mathcal{F}_{B(t)}).
\]
Note that for all saturated arms $j\in C(t)$, we have  
\[
\Delta_j(t) > g(t) s_{j}(B(t)).
\]
In the case that $E^{\mu}(t)$ and $E^{\theta}(t)$ are both true, we have
\[
\theta_j(t)\le \langle b_j(t), \mu \rangle+g(t) s_{j}(B(t)).
\]
 Hence, conditioned on $\mathcal{F}_{B(t)}$ if $E^{\mu}(t)$ is true, we have either the event $E^{\theta}(t)$ is false or for all $j\in C(t)$,
\[
\theta_j(t) \le \langle b_j(t), \mu \rangle+ g(t) s_{j}(B(t)) \le \langle b_{a^*(t)}(t), \mu \rangle ,
\]
Thus, for any $\mathcal{F}_{B(t)}$ that $E^{\mu}(t)$ holds,
\begin{align*}
    \Pr(\theta_{a^*(t)}(t) > \theta_j(t), \forall j\in C(t) | \mathcal{F}_{B(t)}) 
    \ge& \Pr(\theta_{a^*(t)}(t) > \langle b_{a^*(t)}(t), \mu\rangle | \mathcal{F}_{B(t)}) - \Pr(\overline{E^{\theta}(t)}|\mathcal{F}_{B(t)})\\
    \ge&  p-\frac{1}{t^2} \ .
\end{align*}
The above inequalities are due to Lemmas~\ref{lemma:mutheta} and  ~\ref{lem:prteta}. 
\end{proof}

\begin{lemma} \label{lemma:deltafp}
For any filtration $\mathcal{F}_{B(t)}$, assuming $E^{\mu}(t)$ holds true,
\[
\Ex\left[\Delta_{a(t)}(t)|\mathcal{F}_{B(t)}\right] \le \frac{3g(t)}{p} \Ex\left[s_{a(t)}(B(t))|\mathcal{F}_{B(t)}\right]+\frac{2g(t)}{pt^2} \ .
\]
\end{lemma}
\begin{proof}
The proof follows closely \citet[Lemma 4]{contextual} and adapts it to the batch setting. 
First define
\[
\bar{a}(t)=\arg\min_{a \notin C(t)} s_a(B(t)),
\]
Since $\mathcal{F}_{B(t)}$ defines ${\cal B}(B(t))$ and also $b_a(t)$ are independent of unobserved rewards (before making a batch query) thus given $\mathcal{F}_{B(t)}$ and context vectors $b_a(t)$, the value of $\bar{a}(t)$ is determined. Now by applying Lemma~\ref{lemma:pminus},
for any $\mathcal{F}_{B(t)}$ and by assuming that $E^{\mu}(\theta)$ is true, we have
\begin{align*}
    \Ex\left[s_{a(t)}(B(t))|\mathcal{F}_{B(t)}\right] &\ge \Ex\left[s_{a(t)}(B(t))|\mathcal{F}_{B(t)},a(t)\notin C(t)\right] \cdot \Pr(a(t)\notin C(t)|\mathcal{F}_{T-1}) \\
&\ge s_{\bar{a}(t)}(B(t)) (p-\frac{1}{t^2}).
\end{align*}
Again if both $E^{\mu}(t)$ and $E^{\theta}(t)$ are true, then for all $a$ we have,
\[
\theta_{a}(t) \le \langle b_a(t), \mu \rangle+g(t) s_a(B(t)).
\]
Moreover, we know that for all $a$, $\theta_{a(t)}(t)\ge \theta_a(t)$, thus
\begin{align*}
    \Delta_{a(t)}(t)&= \Delta_{\bar{a}(t)}(t)+(\langle b_{\bar{a}(t)}(t), \mu\rangle - \langle b_{a(t)}(t), \mu \rangle)\\
    &\le 2 g(t) s_{\bar{a}(t)}(B(t))+g(t)s_{a(t)}(B(t)).
\end{align*}
Consequently, 
\begin{align*}
\Ex\left[\Delta_{a(t)} | \mathcal{F}_{B(t)}\right] 
&\le \frac{2g(t)}{p-\frac{4}{t^2}} \Ex\left[s_{a(t)}(B(t))|\mathcal{F}_{B(t)}\right] + g(t) \Ex\left[s_{a(t)}(B(t))|\mathcal{F}_{B(t)}\right]+\frac{1}{t^2}\\
&\le \frac{3}{p} g(t) \Ex\left[s_{a(t)}(B(t))|\mathcal{F}_{B(t)}\right] + \frac{2g(t)}{pt^2} \ .
\end{align*}

The first inequality is because $\Delta_a \le 1$ for all $a$.
The second inequality uses Lemma~\ref{lemma:mutheta} to get $\Pr(\overline{E^{\theta}(t)}) \le \frac{1}{t^2}$.
Furthermore, in the last inequality  we use the fact that $0\le s_{a(t)}(B(t)) \le |b_{a(t)}(t)| \le 1$.
\end{proof}
Similar to ~\cite{agrawal2017near} we have the following definitions. 
\begin{definition}
\[
\mathcal{R}'(t):=\mathcal{R}(t) \times \mathbb{I}(E^{\mu}(t)).
\]
\end{definition}
\begin{definition}
Define
\begin{align*}
X_t &= \mathcal{R}'(t)-\frac{3g(t)}{p}  s_{a(t)}(B(t)) - \frac{2g(t)^2}{pt^2 } \\
Y_t&= \sum_{w=1}^{t} X_w.
\end{align*}
\end{definition}

Becasue of the way we defined $Y_t$, namely, the filteration $\mathcal{F}_{B(t)}$, we can easily show the following lemma. 
\begin{lemma}
The sequence $\{Y_t\}_{t=0}^T$ is a super martingale  with respect to $\mathcal{F}_{B(t)}$.
\end{lemma}
\begin{proof}
The proof follows closely \citet[Lemma 5]{contextual} and adapts it to  filtration $\mathcal{F}_{B(t)}$ induced by the batch algorithm. 
Basically, we need to show that for all $t\geq 0$,
\[
\Ex\left[Y_t-Y_{t-1}|\mathcal{F}_{B(t)}\right] \le 0,
\]
In other words
\[
\Ex\left[\mathcal{R}'(t)|\mathcal{F}_{B(t)}\right] \le \frac{3g(t)}{p}\Ex\left[s_{a(B(t))}(t)|\mathcal{F}_{B(t)}\right] + \frac{2g(t)}{pt^2} ,
\]
First, note that $\mathcal{F}_{B(t)}$  determines the event $E^{\mu}(t)$.
Assuming that $\mathcal{F}_{B(t)}$ is such that $E^{\mu}(t)$ is not true, then
$\mathcal{R}'(t)=0$ ] and the above inequality is trivial.
Otherwise, if for $\mathcal{F}_{B(t)}$, the event $E^{\mu}(t)$ holds, Lemma~\ref{lemma:deltafp} implies the result.  
\end{proof}
The following Lemma is a batch variant of \citet[Lemma 3]{chu2011contextual}. 
\begin{lemma}\label{lem:chu}
\begin{align}\label{eq:sums}
\sum_{t=1}^T s_{a(t)}(B(t)) \le 5\sqrt{dT\ln T} \ .
\end{align}
\end{lemma}
\begin{proof}
Upper bounding the expression $\sum_{t} s_{a(t)}(B(t))$  follows by the same steps in  \citet[Lemma 3]{chu2011contextual} for the matrix ${\cal B}(B(t))$ (we loose a constant factor in the process).
The reason is that the term $\sum s_{B(t),a(t)}$ can  be written in terms of eigenvalues of ${\cal B}(B(t))$ matrices. 
More precisely, from Lemma 2 in Chu et al. (2011) we can arrange eigenvalues of ${\cal B}(t)$ to obtain the following  bound $$
s_{a(t)}(B(t))^2 \le 10 \sum_{j} \frac{\lambda_{t+1,j}-\lambda_{t,j}}{\lambda_{t,j}}.$$
Note that the above upper bound is independent of our batch algorithm.
Then for  $\psi=|\Psi_{T+1}|$ (in  \citet[Lemma 3]{chu2011contextual}) 
we have
\begin{align*}
    \sum_{t\in \Psi_{T+1}} s_{a(t)}(B(t)) = \sum_{t\in \Psi_{T+1}} \sqrt{10 \sum_{j}(\frac{\lambda_{t+1,j}}{\lambda_{t,j}}-1)},
\end{align*}
for each matrix ${\cal B}(B(t))$ in $\Psi_{T+1}$. The function $f$ can be defined similar to  \citet[Lemma 3]{chu2011contextual} for $\Psi_{T+1}$.
As in Lemma 3, the ratio of eigenvalues remain greater than or equal 1. The following sum product can  be bounded by $\psi+d$  since the norm of each $x_{ta(t)}$ is bounded by $1$. For $t'={B}(t)$ between $T/2$ and $T+1$,
\begin{align*}
\sum_{j} \prod_{t}  \frac{\lambda_{t+1,j}}{\lambda_{t,j}} \le \sum_{j} \lambda_{t',j} = \sum_{t} ||x_{t,a(t)}||^2 +d \le \psi +d.
\end{align*}
So, we can similarly bound 
$$\sum_{t\in \Psi_{T+1}} s_{a(t)}(B(t)) \le \psi \sqrt{10d}\sqrt{(\psi+1)^{1/\psi}-1}.$$ Thus, by using \citet[Lemma 9]{chu2011contextual} for $\psi$ we can obtain  
eq.~\eqref{eq:sums}.
\end{proof}

\begin{proof}
[{Proof of Theorem~\ref{thm:contextual}}] We rely on the proof technique by \citet[Theorem 1]{contextual}.
 First, note that $X_t$ is bounded as 
 $$|X_t|\le 1+\frac{3}{p}g(t)+\frac{2}{pt^2}g(t)\le \frac{6}{p} g(t).$$
Also $g(t) \le g(T)$. Thus, by applying Azuma-Hoeffding inequality for Martingale sequences, we have
\[
\Pr\left(\sum_{t=1}^T \mathcal{R}'(t) \le  \frac{3g(T)}{p} \sum_{t=1}^T s_{a(t)}(B(t)) + \frac{2g(T)}{p} \sum_{t=1}^T \frac{1}{t^2} + \frac{6g(T)}{p} \sqrt{2T \ln(2/\delta)}\right)\geq  1-\frac{\delta}{2}.
\]
Therefore, by invoking Lemma~\ref{lem:chu} we know that with probability $1-\frac{\delta}{2}$ we have 
\[
\sum_{t=1}^T \mathcal{R}'(t)= O\left(d\sqrt{T}\times (\min\{\sqrt{d},\sqrt{\log N}\})\times (\ln(T)+\sqrt{\ln(T)\ln(1/\delta)})\right).
\]
Furthermore, Lemma~\ref{lemma:mutheta} implies that with probability of at least $1-\delta/2$, the event $E^{\mu}(t)$ holds for all $t$. Thus, with probability of at least $1-\delta$, 
\[
\mathcal{R}(T)=O\left(d\sqrt{T}\times (\min\{\sqrt{d},\sqrt{\log N}\})\times (\ln(T)+\sqrt{\ln(T)\ln(1/\delta)})\right).
\]
\end{proof}